\documentclass[preprint,review,12pt]{elsarticle}

\usepackage{booktabs}
\usepackage{multirow}
\usepackage{amsmath,amssymb}
\usepackage{graphicx}
\usepackage{booktabs}
\usepackage{multirow}
\usepackage{array}
\usepackage{url}
\usepackage{bm}
\usepackage{algorithm}
\usepackage{algorithmic}
\usepackage{lineno}
\usepackage{threeparttable}
\usepackage{placeins}
\usepackage{graphicx}
\usepackage{subcaption}
\usepackage[a4paper,left=0.75in,right=0.75in,top=0.75in,bottom=0.8in]{geometry}

\graphicspath{{figures/}{pdf/figures/}}

\journal{Biomedical Signal Processing and Control}
\begin{document}

\begin{frontmatter}

\title{A Hybrid CNN--State-Space--Attention Backbone with Joint-Embedding Predictive Pretraining for 12-Lead ECG Classification}

\author[ksu1]{Yakoub Bazi\corref{cor1}}
\ead{ybazi@ksu.edu.sa}
\cortext[cor1]{Corresponding author.}

\author[ksu1]{Sarah Aljuhani}
\author[ksu2]{Mohamad M. Al Rahhal}
\author[ksu1]{Mansour Zuair}
\author[ksu1]{Naif Alajlan}

\affiliation[ksu1]{
organization={Computer Engineering Department, College of Computer and Information Sciences, King Saud University},
city={Riyadh},
postcode={11543},
country={Saudi Arabia}
}

\affiliation[ksu2]{
organization={Applied Computer Science Department, College of Applied Computer Science, King Saud University},
city={Riyadh},
postcode={11543},
country={Saudi Arabia}
}

\begin{abstract}
Automatic 12-lead electrocardiogram (ECG) classification requires representations that jointly capture local waveform morphology, long-range temporal dynamics, and cross-lead dependencies, yet integrating these properties within a single efficient architecture remains challenging. This paper introduces a hybrid CNN--SSM--Attention backbone for 12-lead ECG classification. A convolutional stem performs early waveform tokenization and temporal reduction, mixed state-space and depthwise convolutional blocks model temporal dynamics and local morphology, and a late self-attention stage enables global token interaction at reduced resolution. To improve transfer from unlabeled data, we further develop an ECG-oriented Joint-Embedding Predictive Pretraining (JEPA) framework. Unlike ViT-based JEPA methods that mask patch tokens before the encoder, the proposed method samples span masks at the latent temporal resolution and projects them back to the waveform domain, then predicts clean latent targets from a momentum encoder without waveform reconstruction. Experiments on CPSC2018, Chapman-Shaoxing, and PTB-XL, with pretraining on approximately 350K unlabeled CODE-15 recordings, show that the proposed backbone provides strong supervised baselines under a compact parameter budget. JEPA pretraining further improves transfer, particularly in reduced-label settings and under both full fine-tuning and LoRA-based adaptation. Code: \url{https://github.com/yakoubbazi/Hybrid_ECG_Jepa}.
\end{abstract}


\end{frontmatter}

%\linenumbers
\newpage

\section{Introduction}

Electrocardiography (ECG) is one of the most widely used non-invasive tools for cardiac assessment. A standard 12-lead ECG records cardiac electrical activity from complementary anatomical views, making it possible to assess rhythm disorders, conduction abnormalities, ischemic changes, and morphological deviations. Automated 12-lead ECG interpretation, however, remains challenging because clinically relevant patterns appear at multiple temporal and spatial scales. Some diagnostic cues are highly local, such as QRS morphology and ST--T changes, whereas others depend on longer temporal context and consistency across leads \cite{wagner2020ptbxl,perez2020challenge}.

Deep learning has substantially advanced automated ECG classification, but designing an effective backbone for 12-lead signals remains non-trivial. Convolutional neural networks provide strong local inductive biases for waveform morphology and remain competitive ECG baselines. Transformers, on the other hand, enable explicit global interaction among temporal tokens, but their quadratic complexity can be inefficient for long ECG sequences. More recently, selective state-space models (SSMs), including Mamba-style architectures, have provided an efficient alternative for long-context sequence modeling with linear-time complexity \cite{gu2023mamba}. These developments suggest that local morphology extraction, efficient temporal sequence modeling, and global token interaction are complementary rather than mutually exclusive. Nevertheless, scalable hybrid CNN--SSM--Attention designs remain underexplored for 12-lead ECG classification.

A second challenge is the dependence on labeled ECG data. Large collections of raw ECG recordings are increasingly available, but expert annotations remain costly, time-consuming, and often dataset-specific. Self-supervised learning (SSL) has therefore become an important direction for ECG representation learning. Existing methods include contrastive learning, cross-view objectives, masked reconstruction, and large-scale ECG foundation models \cite{kiyasseh2021clocs,mehari2022ssl,huang2023crossdim,ran2023joint,sawano2024mae,ma2026glssl,mckeen2025ecgfm}. Among these directions, joint-embedding predictive architectures (JEPA) are attractive because they avoid direct raw-signal reconstruction and instead learn by predicting latent targets from corrupted inputs. Recent ECG studies indicate that JEPA-style pretraining can learn transferable representations for downstream classification \cite{kim2024ecgjepa,weimann2025jepaecg}.

In this work, we propose a scalable hybrid CNN--SSM--Attention backbone for 12-lead ECG classification. The architecture follows a staged design. A convolutional stem first extracts local waveform morphology and performs early temporal reduction. The intermediate stages use mixed sequence-morphology blocks that combine selective state-space mixing with lightweight depth-wise convolutional processing, allowing the model to capture long-range temporal dependencies while preserving local waveform structure. A final self-attention stage is then applied after temporal compression, enabling explicit global token interaction at a reduced computational cost. This design assigns each modeling mechanism to a suitable part of the network hierarchy: convolution for early local structure, SSM-based mixing for efficient temporal modeling, and attention for high-level global integration.

We further study JEPA-style self-supervised pretraining for the proposed backbone. Unlike ViT-based JEPA formulations, where masking is applied directly to patch tokens before the encoder, our method uses latent-aligned waveform corruption. Specifically, temporal masks are sampled at the latent temporal resolution of the backbone, which is reduced by a factor of four relative to the input waveform. The sampled mask is then expanded back to the raw waveform domain, where the selected ECG regions are replaced by learned lead-wise masking values with small additive noise. The online encoder processes this corrupted ECG, while a momentum target encoder processes the clean signal and provides latent prediction targets. Thus, the method performs masked latent prediction without reconstructing the raw ECG waveform and without inserting explicit ViT-style mask tokens into the encoder.

We evaluate the proposed framework on three public 12-lead ECG benchmarks: CPSC2018, Chapman, and PTB-XL. The experiments include supervised training from random initialization, full fine-tuning from self-supervised initialization, LoRA-based parameter-efficient adaptation, and reduced-label transfer settings. This evaluation is intended to assess both the intrinsic capacity of the proposed backbone and the transfer behavior induced by JEPA-style pretraining.

The main contributions of this work are summarized as follows:
\begin{enumerate}
    \item We propose a scalable hybrid CNN--SSM--Attention backbone for 12-lead ECG classification, integrating convolutional morphology extraction, efficient state-space sequence modeling, and late self-attention within a unified architecture.
    
    \item We introduce an ECG-specific JEPA pretraining strategy with latent-aligned waveform corruption, where masks are sampled at latent resolution and applied in the waveform domain for masked latent prediction.
    
    \item We conduct comprehensive experiments on CPSC2018, Chapman, and PTB-XL, covering supervised training from scratch, full fine-tuning, LoRA-based adaptation, and reduced-label transfer.
\end{enumerate}

The remainder of this paper is organized as follows. Section~\ref{sec:related_work} reviews related work on ECG backbones, state-space sequence modeling, and self-supervised ECG representation learning. Section~\ref{sec:backbone} presents the proposed hybrid ECG backbone. Section~\ref{sec:ssp} describes the JEPA-style self-supervised pretraining framework. Section~\ref{sec:experiments} reports the experimental setup and results. Section~\ref{sec:conclusion} concludes the paper.
\section{Related Work}
\label{sec:related_work}

This section reviews prior work related to the proposed framework. We focus on two directions: backbone design for 12-lead ECG classification and self-supervised ECG representation learning.

\subsection{Backbones for 12-Lead ECG Classification}
\label{sec:relwork_backbones}

Deep learning has become a dominant paradigm for automated 12-lead ECG classification. Early and widely used methods are largely based on one-dimensional convolutional neural networks, which are well matched to the local structure of ECG waveforms. Many diagnostic patterns, including QRS morphology, ST--T abnormalities, and beat-level deformation, are expressed through localized temporal changes. As a result, convolutional models remain strong baselines on public ECG benchmarks such as PTB-XL and CPSC2018 \cite{wagner2020ptbxl,perez2020challenge}. However, ECG interpretation also requires broader temporal reasoning and cross-lead integration, since rhythm abnormalities, conduction patterns, and multilead morphological consistency may depend on information distributed across time and leads.

To address this limitation, recent studies have moved toward hybrid architectures that combine local waveform processing with mechanisms for longer-range or multiscale interaction. DAMS-Net introduces dual attention and multiscale feature fusion to capture complementary local and global representations \cite{zhou2023damsnet}. MSGFormer and MCTNet further explore transformer-style and multiscale token interaction for 12-lead ECG analysis \cite{ji2024msgformer,li2024mctnet}. Other convolution-attention hybrids similarly combine front-end morphology extraction with attention-based aggregation or temporal reasoning modules \cite{bai2024hybrid,msft2024}. These methods suggest that effective ECG backbones should not rely on a single modeling principle, but should instead combine local, temporal, and global representations in a structured manner.

State-space models (SSMs) have recently emerged as an efficient alternative for long-sequence modeling. In a standard discrete formulation, an SSM updates a latent state according to
\begin{equation}
\label{eq:ssm}
\mathbf{s}_{t+1} = \mathbf{A}\mathbf{s}_{t} + \mathbf{B}\mathbf{x}_{t}, \qquad
\mathbf{y}_{t} = \mathbf{C}\mathbf{s}_{t} + \mathbf{D}\mathbf{x}_{t},
\end{equation}
where $\mathbf{x}_{t}$ is the input at time step $t$, $\mathbf{s}_{t}$ is the latent state, and $\mathbf{y}_{t}$ is the output. Selective SSMs, particularly Mamba, make the state update input-dependent and provide linear-time sequence modeling, making them attractive for long ECG recordings \cite{gu2023mamba}. This efficiency is important because clinically relevant ECG dependencies may span multiple beats, while full self-attention can be computationally expensive when applied at high temporal resolution.

Several recent works have adapted Mamba-style architectures to ECG analysis. ECGMamba uses bidirectional Mamba blocks with residual and feed-forward refinement for ECG classification \cite{qiang2024ecgmamba}. ECG-Mamba follows a related direction by building a 12-lead classifier around a bidirectional SSM backbone and task-specific augmentation \cite{jiang2025ecgmamba}. In these approaches, Mamba typically acts as the main temporal modeling component or as a substitute for transformer-style sequence processing.

Different from these works, our goal is not to replace attention with SSMs or to use Mamba as a standalone ECG backbone. Instead, we design a staged hybrid architecture motivated by a distinct modeling consideration. The convolutional stem performs early morphology extraction and temporal reduction. The intermediate stages combine selective state-space mixing with depth-wise convolutional processing to jointly model long-range temporal structure and local waveform patterns. The final stage applies self-attention only after temporal compression, enabling explicit global token interaction with reduced computational cost. This design provides a unified CNN--SSM--Attention hierarchy tailored to 12-lead ECG signals.

\subsection{Self-Supervised ECG Representation Learning}
\label{sec:relwork_ssl}

A major challenge in ECG learning is the limited availability of high-quality diagnostic labels, which has motivated a broad range of SSL methods. Early ECG SSL approaches were dominated by contrastive learning. Representative examples include CLOCS, which constructs contrastive objectives across temporal, spatial, and patient-level views \cite{kiyasseh2021clocs}, and later multilead ECG SSL methods that exploit ECG-specific augmentations or cross-view consistency \cite{mehari2022ssl,huang2023crossdim,ran2023joint,liu2023denselead,liu2024morphology,liu2025lcd}. A parallel line of work explores masked reconstruction or masked autoencoding, where the model reconstructs either raw waveforms or latent features from corrupted inputs \cite{sawano2024mae,mckeen2025ecgfm,shi2024universal}.

More recently, JEPA-style latent predictive learning has emerged as an alternative to both contrastive learning and raw-signal reconstruction. Instead of enforcing invariance across augmented views or reconstructing waveform samples directly, JEPA predicts target representations in latent space using a momentum target encoder \cite{assran2023ijepa}. This formulation is particularly appealing for ECG because the downstream objective is not faithful recovery of every low-level fluctuation, but the learning of transferable latent structure that captures waveform morphology, rhythm patterns, and inter-lead relationships.

Recent ECG studies have started to explore this direction explicitly. ECG-JEPA adapts JEPA to 12-lead ECGs using masked latent prediction and a transformer-based encoder, and further introduces a cross-pattern attention design tailored to multilead ECG structure \cite{kim2024ecgjepa}. In another work, the authors also study JEPA-based self-supervised pretraining for ECG classification and report that latent predictive learning can outperform both invariance-based and generative pretraining strategies on PTB-XL benchmarks \cite{weimann2025jepaecg}. These results indicate that JEPA is a promising self-supervised paradigm for ECG representation learning.

Our pretraining strategy follows this JEPA direction but is paired with a different backbone design. Unlike prior ECG-JEPA work built primarily on transformer-style encoders, the present work couples JEPA with the proposed hybrid CNN--SSM--Attention backbone. Pretraining is based on masked latent prediction with a momentum target encoder and ECG-specific structured corruption, without waveform reconstruction or contrastive pairing. In this way, JEPA serves as a representation-learning strategy that complements the proposed backbone rather than defining the main architectural contribution.

\subsection{Relation to the proposed method}

The above literature suggests two main trends. First, 12-lead ECG backbones increasingly benefit from combining local waveform modeling with mechanisms that capture broader temporal and inter-lead context. Second, self-supervised ECG learning is moving from generic contrastive objectives toward masked and latent predictive formulations that better match the structure of multilead signals.

The proposed method combines these two directions in a unified framework. Its main contribution is a scalable hybrid CNN--SSM--Attention backbone in which convolution, state-space mixing, and self-attention are assigned distinct roles across the network hierarchy. JEPA-style pretraining is then used as a complementary self-supervised strategy for learning transferable latent representations on top of this backbone. In this sense, the proposed method lies at the intersection of recent hybrid ECG backbone design, emerging SSM-based ECG sequence modeling, and latent predictive SSL for multilead ECGs.

\section{Proposed Scalable Hybrid ECG Backbone}
\label{sec:backbone}

\subsection{Overview}
\label{sec:method_overview}

We propose a scalable hybrid backbone for 12-lead ECG classification. The backbone is organized as a three-stage hierarchy that progressively transforms the raw multi-lead signal into a compact token representation. The overall design is guided by a simple idea: local waveform morphology should be captured early, longer temporal dependencies should be modeled efficiently as the sequence becomes shorter, and self-attention should be introduced only in the final stage, where global interaction is more affordable.

Let the input ECG segment be denoted by
\begin{equation}
\mathbf{X}\in\mathbb{R}^{C\times T},
\end{equation}
where $C=12$ is the number of leads and $T$ is the temporal length. In our setting, each ECG is represented as a 10-second recording sampled at 100~Hz, giving $T=1000$ time samples. The backbone maps $\mathbf{X}$ to a latent token sequence
\begin{equation}
\mathbf{H}=g_{\theta}(\mathbf{X})\in\mathbb{R}^{T'\times D},
\end{equation}
where $T'$ is the final token length and $D$ is the output feature dimension. In our implementation, the temporal resolution is reduced by a factor of four, yielding a final sequence length of approximately $T/4$.

\begin{figure*}[!t]
    \centering
    \includegraphics[width=\textwidth]{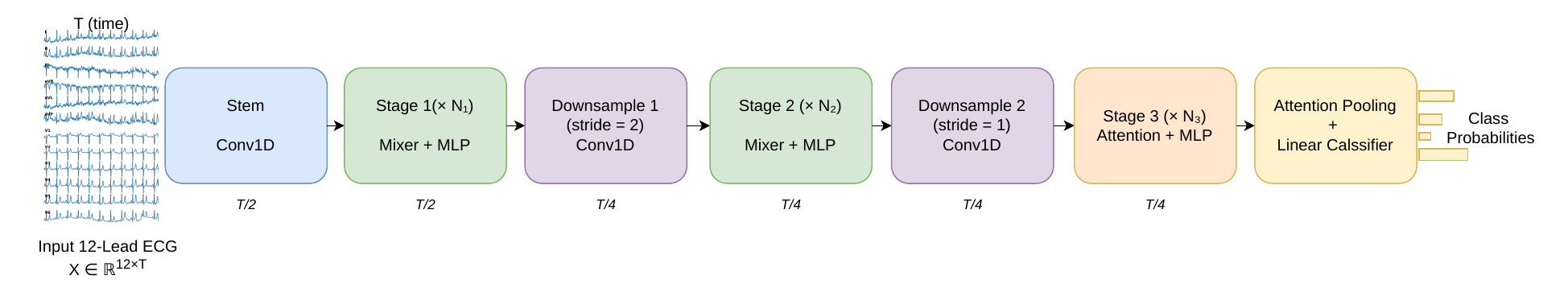}
    \caption{Overview of the proposed hybrid CNN--SSM--Attention backbone for 12-lead ECG classification. The architecture combines early convolutional tokenization, hybrid mixer stages, late self-attention, and attention pooling for final classification.}
    \label{fig:backbone_overall}
\end{figure*}

As shown in Fig.~\ref{fig:backbone_overall}, the network begins with a convolutional stem that performs early lead fusion and initial temporal downsampling. Stage~1 and Stage~2 then apply hybrid mixer blocks that combine efficient temporal modeling with local convolutional refinement. Stage~3 operates on the compressed sequence and uses self-attention to capture global relationships across tokens. Finally, the resulting token sequence is aggregated by attention pooling and passed to a classifier to produce the diagnostic prediction.

\subsection{Convolutional Stem}
The convolutional stem performs early lead fusion and temporal downsampling with a 1-D convolution of kernel size $7$, stride $2$, and padding $3$, followed by group normalization and a SiLU nonlinearity. The kernel size of $7$ provides a moderately broad temporal receptive field for early waveform extraction while keeping the stem lightweight:
\begin{equation}
\mathbf{H}_1
=
\mathrm{SiLU}\!\bigl(
\mathrm{GN}(\mathrm{Conv1D}(\mathbf{X}))
\bigr)
\in \mathbb{R}^{T_1 \times C_1},
\label{eq:stem}
\end{equation}
where $T_1 = T/2$ denotes the reduced temporal length and $C_1$ denotes the stem feature dimension. This stage jointly integrates cross-lead information and reduces the temporal resolution, producing a compact representation for the subsequent backbone stages.

\begin{figure}[t]
    \centering
    \includegraphics[width=\linewidth]{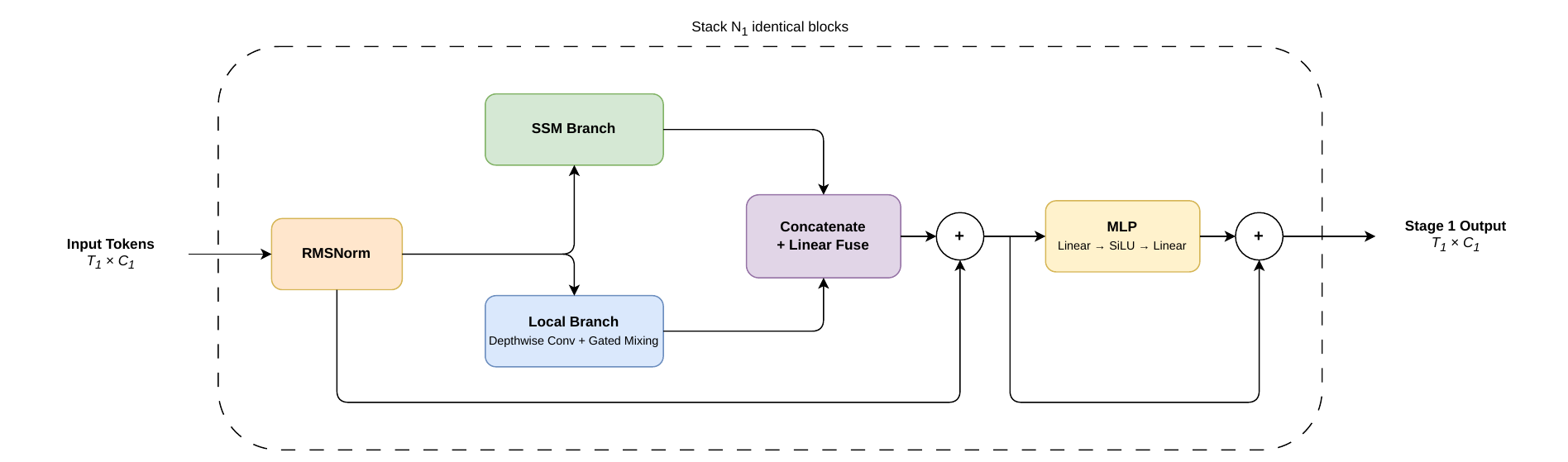}
   \caption{Architecture of a Stage~1 block. The block operates at resolution $T_1 \times C_1$ and combines an SSM branch for temporal sequence modeling with a local gated depthwise-convolution branch for morphology-aware refinement, followed by residual fusion and an MLP sublayer.}
    \label{fig:stage1_block}
\end{figure}

\subsection{Stage 1: Hybrid mixer-based sequence modeling}
Stage 1 takes the stem output $\mathbf{H}_1$ as input and applies $N_1$ repeated blocks, each consisting of a hybrid mixer followed by an MLP. As shown in Fig.~\ref{fig:stage1_block}, the hybrid mixer combines an SSM branch for temporal sequence modeling with a local gated depthwise-convolution branch for morphology-aware refinement. The rationale is to couple efficient long-range temporal mixing with explicit local waveform modeling within the same block. In particular, the SSM branch captures broader temporal dependencies, while the local branch is intended to preserve short-range morphological structure. This design is inspired by the MambaVision mixer proposed for visual backbones~\cite{hatamizadeh2025mambavision}, where a non-SSM convolutional path is introduced to complement the sequential branch and enrich the resulting representation, but is adapted here to 1-D multi-lead ECG signals. After fusion of the two branches, an MLP further refines the representation.

The computation within a single Stage~1 block can be described as follows. Given an input sequence $\mathbf{H}_1\in\mathbb{R}^{T_1\times C_1}$, the block first applies RMS normalization:
\begin{equation}
\widetilde{\mathbf{H}}_1=\mathrm{RMSNorm}(\mathbf{H}_1).
\end{equation}
The first branch then applies a state-space sequence modeling module:
\begin{equation}
\mathbf{A}=\mathrm{SSM}(\widetilde{\mathbf{H}}_1),
\label{eq:ssm_branch}
\end{equation}
which serves as an efficient temporal mixing operator for capturing longer-range dependencies in the ECG sequence~\cite{gu2023mamba}.

The second branch applies depthwise convolution followed by gated channel mixing:
\begin{equation}
\mathbf{U},\mathbf{V}
=
\mathrm{split}\!\Bigl(
W_{\mathrm{in}}\,\mathrm{DWConv}(\widetilde{\mathbf{H}}_1)
\Bigr),
\end{equation}
\begin{equation}
\mathbf{B}
=
W_{\mathrm{out}}\!\Bigl(
\mathrm{SiLU}(\mathbf{U})\odot\mathbf{V}
\Bigr),
\label{eq:local_branch}
\end{equation}
where $\mathrm{DWConv}(\cdot)$ denotes depthwise 1-D convolution, $\odot$ denotes element-wise multiplication, and $W_{\mathrm{in}}$ and $W_{\mathrm{out}}$ are learned input and output projection matrices of the local branch. The depthwise convolution provides lightweight local temporal filtering, while the gating operation adaptively modulates the resulting features so that informative short-range waveform patterns can be emphasized before fusion.

The outputs of the two branches are then concatenated, linearly fused, and added through a residual connection:
\begin{equation}
\mathbf{H}_1'
=
\mathbf{H}_1
+
\mathrm{Dropout}\!\Bigl(
W_{\mathrm{fuse}}[\mathbf{A};\mathbf{B}]
\Bigr),
\label{eq:mixer}
\end{equation}
thereby retaining complementary information from the SSM and local convolutional paths within a unified token representation.

The mixer output $\mathbf{H}_1'$ is then passed to the MLP, which applies two successive linear projections separated by a SiLU nonlinearity:
\begin{equation}
\begin{aligned}
\mathrm{MLP}(\mathbf{H}_1')
&=
\mathbf{H}_1'
+
\mathrm{Dropout}\!\Bigl(
W_{\mathrm{mlp},2}\,
\mathrm{SiLU}(\mathbf{Q})
\Bigr),
\\
\mathbf{Q}
&=
W_{\mathrm{mlp},1}\,
\mathrm{RMSNorm}(\mathbf{H}_1').
\end{aligned}
\label{eq:ffn}
\end{equation}
The MLP further refines the fused representation through channel mixing by first expanding the feature dimension with $W_{\mathrm{mlp},1}$ and then projecting it back to the original dimension with $W_{\mathrm{mlp},2}$. Repeating this block $N_1$ times yields the Stage~1 output, denoted by $\mathbf{Z}_1\in\mathbb{R}^{T_1\times C_1}$, while preserving the feature resolution.

\subsection{Stage 2: Hybrid mixer-based sequence modeling at reduced resolution}
Stage 2 begins with a convolutional transition applied to the Stage~1 output $\mathbf{Z}_1$, which further reduces the temporal resolution and expands the feature dimension:
\begin{equation}
\mathbf{H}_2
=
\mathrm{Conv1D}(\mathbf{Z}_1)
\in \mathbb{R}^{T_2 \times C_2},
\label{eq:down1}
\end{equation}
where the convolution uses kernel size $5$, stride $2$, and padding $2$, yielding $T_2 = T/4$, while $C_2$ denotes the expanded feature dimension. The resulting sequence is then processed by the same hybrid mixer-based blocks used in Stage~1. Accordingly, Stage~2 retains the same combination of SSM-based temporal mixing and local gated depthwise-convolutional refinement, but applies it to a shorter and higher-dimensional representation. This allows the model to capture longer-range temporal structure more efficiently while preserving sensitivity to local waveform morphology. Repeating these blocks yields the Stage~2 output, denoted by $\mathbf{Z}_2\in\mathbb{R}^{T_2\times C_2}$.

\subsection{Stage 3: Global sequence modeling with self-attention}
\begin{figure}[t]
    \centering
    \includegraphics[width=\linewidth]{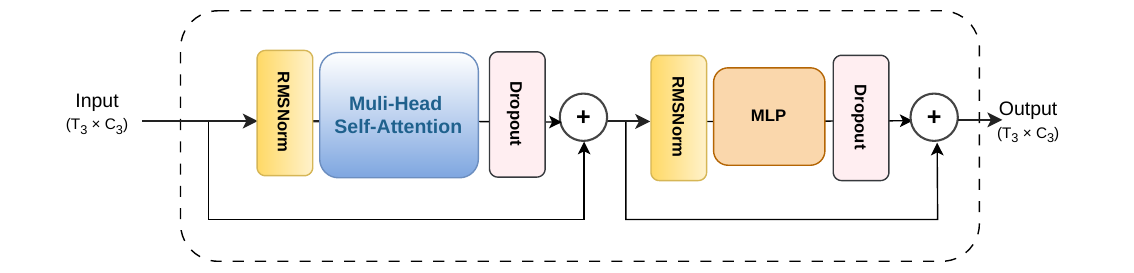}
    \caption{Architecture of a Stage~3 block. Operating at resolution $T_3 \times C_3$, each block follows a Transformer-style design composed of a multi-head self-attention sublayer and an MLP sublayer, with a residual connection around each component.}
    \label{fig:stage3_block}
\end{figure}

Stage 3 begins with a convolutional transition applied to the Stage~2 output $\mathbf{Z}_2$, which projects the representation to a higher-dimensional space while preserving the temporal resolution:
\begin{equation}
\mathbf{H}_3
=
\mathrm{Conv1D}(\mathbf{Z}_2)
\in\mathbb{R}^{T_3\times C_3},
\label{eq:down2}
\end{equation}
where the convolution uses kernel size $5$, stride $1$, and padding $2$, yielding $T_3 = T_2$, while $C_3$ denotes the expanded feature dimension. Since the temporal resolution has already been reduced before Stage~3, no further downsampling is applied at this point; instead, the transition preserves the compressed token sequence while projecting it to a higher-dimensional space prior to global self-attention. Unlike the preceding stages, Stage~3 adopts a pure Transformer-style design composed of repeated self-attention and MLP blocks, as illustrated in Fig.~\ref{fig:stage3_block}. Operating at the reduced temporal resolution makes self-attention computationally affordable while enabling explicit global interaction across tokens. In this way, Stage~3 complements the earlier SSM-based stages by emphasizing long-range dependencies and global contextual integration at the highest representation level.

The computation within a single Stage~3 block can be described as follows. Given an input sequence $\mathbf{H}_3\in\mathbb{R}^{T_3\times C_3}$, the self-attention sublayer is defined as
\begin{equation}
\mathbf{H}_3'
=
\mathbf{H}_3
+
\mathrm{Dropout}\!\Bigl(
\mathrm{MHSA}\!\bigl(
\mathrm{RMSNorm}(\mathbf{H}_3)
\bigr)
\Bigr),
\label{eq:attn}
\end{equation}
where $\mathrm{MHSA}$ denotes multi-head self-attention. The resulting representation is then refined by an MLP sublayer:
\begin{equation}
\begin{aligned}
\mathrm{MLP}(\mathbf{H}_3')
&=
\mathbf{H}_3'
+
\mathrm{Dropout}\!\Bigl(
W_{\mathrm{mlp},2}\,
\mathrm{SiLU}(\mathbf{Q})
\Bigr),
\\
\mathbf{Q}
&=
W_{\mathrm{mlp},1}\,
\mathrm{RMSNorm}(\mathbf{H}_3').
\end{aligned}
\label{eq:stage3_mlp}
\end{equation}
Thus, each Stage~3 block follows the standard pre-normalized Transformer pattern of self-attention followed by an MLP, with a residual connection around each sublayer. Repeating this block $N_3$ times yields the final encoded representation, denoted by $\mathbf{Z}_3\in\mathbb{R}^{T_3\times C_3}$.

\subsection{Classification head}
To obtain a compact sequence-level representation, the final encoded sequence produced by Stage~3, denoted by $\mathbf{Z}_3=\{\mathbf{z}_t\}_{t=1}^{T_3}\in\mathbb{R}^{T_3\times C_3}$, is aggregated using attention pooling. This choice allows the model to assign larger weights to diagnostically informative temporal regions, rather than treating all tokens equally as in average pooling. The attention weight assigned to each token is computed as
\begin{equation}
\alpha_t=
\frac{
\exp\!\left(
\mathbf{w}^{\top}\tanh(\mathbf{W}_p\mathbf{z}_t)
\right)
}{
\sum_{t'=1}^{T_3}
\exp\!\left(
\mathbf{w}^{\top}\tanh(\mathbf{W}_p\mathbf{z}_{t'})
\right)
},
\label{eq:attn_pool_weights}
\end{equation}
where $\mathbf{W}_p$ and $\mathbf{w}$ are learnable parameters of the attention-pooling module. The pooled representation is then given by
\begin{equation}
\bar{\mathbf{z}}=\sum_{t=1}^{T_3}\alpha_t\mathbf{z}_t.
\label{eq:attn_pool}
\end{equation}
The aggregated feature is finally passed to a linear classifier:
\begin{equation}
\hat{\mathbf{p}}=
\mathrm{softmax}(\mathbf{W}_c\bar{\mathbf{z}}+\mathbf{b}_c),
\label{eq:classifier}
\end{equation}
where $\mathbf{W}_c$ and $\mathbf{b}_c$ denote the classifier weight matrix and bias vector, respectively.

\begin{table}[t]
\centering
\caption{Scalable backbone configurations used in this work. $C_1$, $C_2$, and $C_3$ are the channel dimensions of Stages~1--3, and $N_1$, $N_2$, and $N_3$ are the numbers of blocks in each stage. Heads is the number of attention heads in the Stage~3 self-attention blocks, and Params is the total number of parameters.}
\label{tab:model_configs}
\renewcommand{\arraystretch}{1.1}
\setlength{\tabcolsep}{5pt}
\scriptsize
\begin{tabular}{lcccccccc}
\toprule
\textbf{Variant} & $C_1$ & $C_2$ & $C_3$ & $N_1$ & $N_2$ & $N_3$ & \textbf{Heads} & \textbf{Params} \\
\midrule
Micro & 32 & 64  & 64  & 1 & 1 & 1 & 2 & 0.21M \\
Mini  & 48 & 96  & 96  & 1 & 2 & 2 & 2 & 0.78M \\
Tiny  & 64 & 128 & 256 & 2 & 2 & 2 & 4 & 3.11M \\
Small & 64 & 128 & 256 & 2 & 2 & 3 & 4 & 4.17M \\
\bottomrule
\end{tabular}
\end{table}

\subsection{Scalable model variants}
We implement the backbone in four sizes: Micro, Mini, Tiny, and Small (Table~\ref{tab:model_configs}). All variants share the same three-stage design and differ only in channel width, stage depth, and number of attention heads. Scaling capacity through width and depth, while keeping the architecture fixed, is common practice in hierarchical backbones and lets us study the effect of model size without changing the underlying structure.

This scalability serves two purposes. It lets the same design fit different computational budgets, from lightweight to more expressive models, and it enables a controlled study of how representation quality changes with capacity under a fixed CNN--SSM--Attention formulation. The models remain compact overall, ranging from 0.21M parameters for Micro to 4.17M for Small, which is lightweight relative to typical transformer-based ECG backbones. Unless stated otherwise, we use the Small variant for the main pretraining and downstream experiments.

\section{Self-Supervised Pretraining}
\label{sec:ssp}

We pretrain the proposed backbone on unlabeled 12-lead ECG recordings using a joint-embedding predictive objective inspired by JEPA~\cite{assran2023ijepa}. The model is not trained to reconstruct the raw ECG waveform. Instead, it predicts the latent representation of masked temporal regions from a corrupted multilead input. This design encourages the encoder to learn waveform representations that preserve local morphology, temporal context, and cross-lead structure in feature space.

\begin{figure*}[!t]
    \centering
    \includegraphics[width=0.95\textwidth]{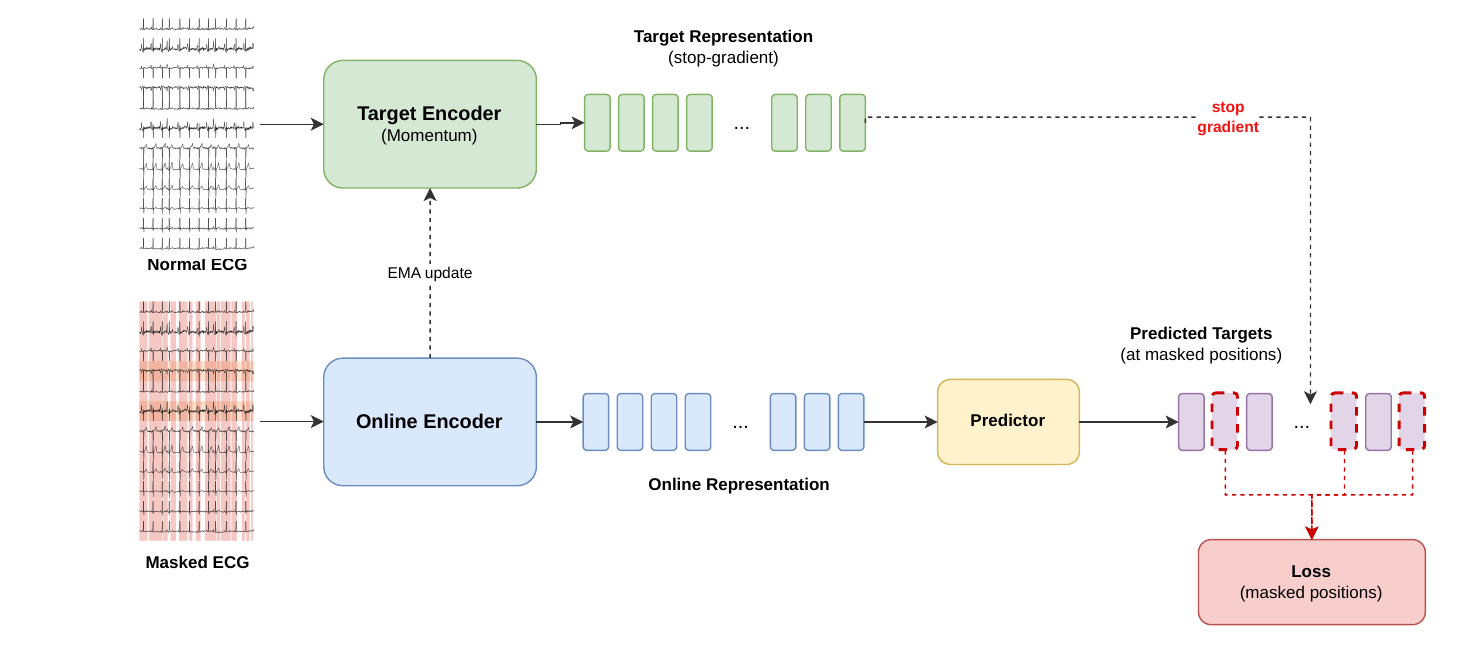}
    \caption{Overview of the proposed ECG-JEPA pretraining scheme. A clean 12-lead ECG segment is processed by the momentum target encoder \(f_{\xi}\) to produce the final Stage~3 target representation \(\mathbf{Z}_3\). A corrupted view of the same signal is processed by the online encoder \(f_{\theta}\), followed by the predictor \(g_{\phi}\), to predict masked final-stage latent targets. The loss is computed only over masked temporal positions.}
    \label{fig:ecg_jepa_pretraining}
\end{figure*}

As illustrated in Fig.~\ref{fig:ecg_jepa_pretraining}, the pretraining framework consists of an online encoder \(f_{\theta}\), a momentum target encoder \(f_{\xi}\), and a predictor \(g_{\phi}\). Given a clean ECG segment
\(\mathbf{X}\in\mathbb{R}^{12\times T}\), a corrupted view \(\tilde{\mathbf{X}}\) is generated before the online encoder. The clean signal \(\mathbf{X}\) is processed by the momentum target encoder, while the corrupted signal \(\tilde{\mathbf{X}}\) is processed by the online encoder. The target encoder is updated as an exponential moving average of the online encoder:
\begin{equation}
\xi \leftarrow m\xi + (1-m)\theta,
\label{eq:ema}
\end{equation}
where \(\theta\) and \(\xi\) denote the online and target encoder parameters, respectively, and \(m\) is the momentum coefficient.

Both encoders produce final-stage temporal representations. The target encoder maps the clean ECG to
\[
\mathbf{Z}_3
=
f_{\xi}(\mathbf{X})
=
\{\mathbf{z}_t\}_{t=1}^{T_3}
\in\mathbb{R}^{T_3\times C_3},
\]
where \(T_3\) is the final temporal token length and \(C_3\) is the feature dimension. In parallel, the online encoder maps the corrupted ECG to
\[
\tilde{\mathbf{Z}}_3
=
f_{\theta}(\tilde{\mathbf{X}})
=
\{\tilde{\mathbf{z}}_t\}_{t=1}^{T_3}
\in\mathbb{R}^{T_3\times C_3}.
\]
The prediction target is defined at the final Stage~3 representation level, which is the same representation later used by the downstream attention-pooling classifier.

The predictor \(g_{\phi}\) maps the online representation \(\tilde{\mathbf{Z}}_3\) to the target latent space. The loss is computed only over the masked temporal token set \(\mathcal{M}\):
\begin{equation}
\mathcal{L}_{\mathrm{JEPA}}
=
\frac{1}{|\mathcal{M}|}
\sum_{t\in\mathcal{M}}
\mathrm{SmoothL1}
\left(
g_{\phi}(\tilde{\mathbf{Z}}_3)_t,
\mathrm{sg}(\mathbf{z}_t)
\right),
\label{eq:jepa_loss}
\end{equation}
where \(g_{\phi}(\tilde{\mathbf{Z}}_3)_t\) is the predicted latent vector at masked position \(t\), \(\mathbf{z}_t\) is the corresponding target vector from the momentum encoder, and \(\mathrm{sg}(\cdot)\) denotes stop-gradient. No waveform reconstruction loss or auxiliary representation loss is used.

The corruption process is defined by two operations. First, random temporal span masks are sampled at the final latent temporal resolution with an overall masking ratio of 75\%. Each selected token corresponds to a temporal region in the input signal; therefore, the sampled mask is expanded back to the waveform domain and the corresponding raw ECG regions are corrupted before entering the online encoder. This differs from ViT-style masked modeling, where explicit mask tokens are inserted into the token sequence. In our case, the encoder receives a corrupted waveform rather than learnable mask-token embeddings. Second, lead dropout is applied with probability 0.25 by selecting one or two complete leads and replacing them with noise. This simulates missing or unreliable channels and encourages the model to exploit cross-lead redundancy.

Overall, the proposed pretraining scheme learns by predicting clean final-stage latent targets from corrupted multilead ECG inputs. By aligning the predictor output with \(\mathbf{Z}_3\), the objective directly optimizes the representation level used for downstream classification.
\subsection{LoRA-Based Adaptation}
\label{sec:lora_adaptation}

In addition to full fine-tuning, we evaluate parameter-efficient adaptation using LoRA. For a pretrained linear layer with frozen weight
$\mathbf{W}_0\in\mathbb{R}^{d_{\mathrm{out}}\times d_{\mathrm{in}}}$, LoRA introduces a trainable low-rank update:
\begin{equation}
\mathbf{W}_{\mathrm{LoRA}}
=
\mathbf{W}_0
+
\frac{\alpha}{r}\mathbf{B}\mathbf{A},
\label{eq:lora}
\end{equation}
where $\mathbf{A}\in\mathbb{R}^{r\times d_{\mathrm{in}}}$ and $\mathbf{B}\in\mathbb{R}^{d_{\mathrm{out}}\times r}$ are trainable matrices, $r$ is the adapter rank, and $\alpha$ is the scaling factor. In our backbone, LoRA is applied to selected linear projections inside the encoder, including attention and MLP-related linear layers. During LoRA adaptation, the original pretrained weights $\mathbf{W}_0$ remain frozen, while only the low-rank adapter parameters are updated. The attention-pooling layer and classification head remain fully trainable.

This setting provides a parameter-efficient way to adapt the SSL-pretrained ECG encoder to downstream datasets. It reduces the number of trainable encoder parameters, helps preserve the pretrained representation, and can be useful when labeled ECG data are limited. In our experiments, LoRA uses rank $r=32$, scaling factor $\alpha=128$, and dropout 0.05.

\section{Experiments}
\label{sec:experiments}

This section presents the experimental evaluation of the proposed framework. We first describe the experimental setup, including datasets, preprocessing, implementation details, and evaluation protocol. We then evaluate the backbone variants under supervised training from scratch, followed by JEPA-style transfer under full-label and reduced-label settings. Finally, we compare the proposed method with existing 12-lead ECG classification approaches.

\subsection{Datasets and preprocessing}

We use four public 12-lead ECG datasets in this study. CODE-15\% is used for self-supervised pretraining, whereas CPSC2018, Chapman-Shaoxing, and PTB-XL are used for downstream evaluation. CODE-15\% contains 345,779 ECG records from 233,770 patients and is used only as an unlabeled waveform source. CPSC2018 contains 6,877 recordings sampled at 500~Hz with durations between 6 and 60~s. Chapman-Shaoxing contains 10,646 10-second recordings sampled at 500~Hz. PTB-XL contains 21,837 10-second recordings from 18,885 subjects and is provided at both 100~Hz and 500~Hz.

For downstream evaluation, CPSC2018 is formulated as a 9-class single-label task with \emph{Normal/SNR} (normal sinus rhythm), \emph{AF} (atrial fibrillation), \emph{IAVB} (first-degree atrioventricular block), \emph{LBBB} (left bundle branch block), \emph{RBBB} (right bundle branch block), \emph{PAC} (premature atrial contraction), \emph{PVC} (premature ventricular contraction), \emph{STD} (ST-segment depression), and \emph{STE} (ST-segment elevation). Chapman--Shaoxing is evaluated under a 4-class rhythm setting with \emph{AFIB} (atrial fibrillation), \emph{GSVT} (general supraventricular tachycardia), \emph{SB} (sinus bradycardia), and \emph{SR} (sinus rhythm). PTB-XL follows the diagnostic superclass protocol with \emph{NORM} (normal ECG), \emph{MI} (myocardial infarction), \emph{STTC} (ST/T change), \emph{CD} (conduction disturbance), and \emph{HYP} (hypertrophy).

For data splitting, CPSC2018 follows a fold-based protocol with folds 1--8 for training, fold 9 for validation, and fold 10 for testing. Chapman-Shaoxing uses a stratified 80/10/10 train/validation/test split. PTB-XL preserves the official patient-wise split, using folds 1--8 for training, fold 9 for validation, and fold 10 for testing. The resulting split statistics after single-label filtering are summarized in Table~\ref{tab:dataset_statistics}. The train sets remain imbalanced, with maximum-to-minimum class ratios of 8.58, 5.38, and 17.45 for CPSC2018, Chapman-Shaoxing, and PTB-XL, respectively.

\begin{table}[t]
\centering
\caption{Downstream dataset statistics after single-label filtering. Majority/minority classes and Max/Min ratios are computed from the training split.}
\label{tab:dataset_statistics}
\vspace{2pt}
\renewcommand{\arraystretch}{1.05}
\setlength{\tabcolsep}{3.0pt}
\footnotesize
\begin{tabular}{lccccccc}
\toprule
\textbf{Dataset} & \textbf{Classes} & \textbf{Train} & \textbf{Val} & \textbf{Test} & \textbf{Majority} & \textbf{Minority} & \textbf{Max/Min} \\
\midrule
CPSC2018 & 9 & 5,121 & 640 & 640 & RBBB (1,227) & LBBB (143) & 8.58 \\
Chapman-Shaoxing & 4 & 6,871 & 859 & 859 & SB (3,085) & GSVT (573) & 5.38 \\
PTB-XL & 5 & 12,957 & 1,637 & 1,650 & NORM (7,243) & HYP (415) & 17.45 \\
\bottomrule
\end{tabular}
\end{table}

All datasets are converted to a unified input format. Signals are reordered to the canonical lead arrangement
$\{\mathrm{I}, \mathrm{II}, \mathrm{III}, \mathrm{aVR}, \mathrm{aVL}, \mathrm{aVF}, \mathrm{V1}, \mathrm{V2}, \mathrm{V3}, \mathrm{V4}, \mathrm{V5}, \mathrm{V6}\}$.
Recordings are resampled to 100~Hz and represented as fixed-length 10-second segments, yielding $12 \times 1000$ waveform tensors. Longer recordings are center-cropped, whereas shorter recordings are resized to the target length using linear interpolation. During training, a zero-phase fourth-order Butterworth band-pass filter is applied before per-lead normalization.

\subsection{Implementation details}

All models are implemented in PyTorch and trained with fixed-size $12 \times 1000$ ECG inputs. For self-supervised pretraining, the backbone is trained on 345,779 unlabeled CODE-15\% recordings. We use AdamW with learning rate $7.4\times10^{-5}$, weight decay 0.05, and $(\beta_1,\beta_2)=(0.9,0.95)$. Training is performed for 50K optimization steps with batch size 16 and gradient accumulation over 8 steps, resulting in an effective batch size of 128. The learning rate is linearly warmed up for 4K steps and then decayed with a cosine schedule using a minimum learning-rate ratio of 0.01. Gradients are clipped to a maximum norm of 1.0. The momentum target encoder is updated by exponential moving average, with the decay coefficient scheduled from 0.998 to 0.9999.

The self-supervised objective is masked latent prediction with a Smooth-$\ell_1$ loss $(\beta=1.0)$. The predictor is a transformer-style module with four blocks, four attention heads, dropout 0.1, and hidden dimension $d_p=\max(1.5C_3,128)$, where $C_3$ denotes the final encoder dimension. For the Small backbone, $C_3=256$ and $d_p=384$. The online branch receives the corrupted ECG signal, whereas the momentum branch processes the clean signal and provides stop-gradient latent targets.

The masking policy is fixed during pretraining. Random temporal spans are sampled at the latent temporal resolution using a masking ratio of 75\% and span lengths of 120--400~ms. The sampled masks are expanded back to the waveform domain, where the corresponding ECG regions are replaced by learned lead-wise masking values with small additive noise. Lead dropout is applied with probability 0.25 by replacing one or two complete leads with noise.

For downstream classification, an attention-pooling layer and a lightweight classification head are attached to the encoder. Models are trained for up to 50 epochs using AdamW, weight decay 0.05, cosine learning-rate decay, and a 5\% warm-up ratio. We use batch size 16 with gradient accumulation over 2 steps. The checkpoint with the highest validation macro-F1 is selected for testing, and early stopping is applied. Class imbalance is handled using class-reweighted cross-entropy with inverse-square-root frequency weighting.

We consider three training modes. In scratch training, all parameters are optimized from random initialization. In full SSL fine-tuning, the pretrained encoder, attention-pooling layer, and classification head are updated jointly. In LoRA-based SSL adaptation, the pretrained encoder is frozen except for low-rank adapters inserted into selected linear layers, while the attention-pooling layer and classifier remain trainable. LoRA uses rank $r=32$, scaling factor $\alpha=128$, and dropout 0.05. The learning rates are $1\times10^{-4}$ for scratch training, $3\times10^{-5}$ for full SSL fine-tuning, and $1\times10^{-4}$ for LoRA-based SSL adaptation. All experiments are conducted on a single NVIDIA RTX A6000 GPU with 48 GB of memory.

\subsection{Evaluation protocol}

We report AUC, accuracy, macro-F1, and Cohen's $\kappa$. Macro-F1 is treated as the primary metric because the downstream datasets contain imbalanced diagnostic or rhythm categories. AUC, accuracy, and $\kappa$ are reported as complementary measures of class separability, overall correctness, and agreement beyond chance.

For each downstream run, the checkpoint with the highest validation macro-F1 is selected for final test evaluation. In the reduced-label setting, we sample stratified subsets of the labeled training set using 1\%, 5\%, 10\%, and 20\% of the available training data. For each fraction, the same subset manifest is shared across scratch, SSL-Full, and SSL-LoRA to ensure matched comparisons. Each reduced-label experiment is repeated three times with different sampling seeds, and results are reported as mean $\pm$ standard deviation.

\subsection{Supervised Baselines from Scratch}
\label{sec:scratch_results}
We first train the backbone family from random initialization, without any self-supervised pretraining. The goal is to see how model size affects performance before we add self-supervised learning. We test the four proposed variants, Micro, Mini, Tiny, and Small, which grow in size from 0.21M to 4.17M parameters and differ in channel width and the number of blocks.

\begin{table*}[t]
\centering
\caption{Scratch performance of the backbone variants on CPSC2018, Chapman-Shaoxing, and PTB-XL. All values are reported in percentage (\%).}
\label{tab:scratch_results}
\vspace{2pt}
\renewcommand{\arraystretch}{1.05}
\setlength{\tabcolsep}{3.2pt}
\footnotesize
\begin{tabular}{lcccccccccccc}
\toprule
\multirow{2}{*}{\textbf{Variant}}
& \multicolumn{4}{c}{\textbf{CPSC2018}}
& \multicolumn{4}{c}{\textbf{Chapman-Shaoxing}}
& \multicolumn{4}{c}{\textbf{PTB-XL}} \\
\cmidrule(lr){2-5} \cmidrule(lr){6-9} \cmidrule(lr){10-13}
& \textbf{AUC} & \textbf{Acc} & \textbf{F1} & \textbf{$\kappa$}
& \textbf{AUC} & \textbf{Acc} & \textbf{F1} & \textbf{$\kappa$}
& \textbf{AUC} & \textbf{Acc} & \textbf{F1} & \textbf{$\kappa$} \\
\midrule
Micro 
& 94.66 & 76.88 & 71.58 & 72.90
& \textbf{98.60} & 93.60 & 91.71 & 90.64
& \textbf{90.02} & 76.97 & 62.13 & 62.99 \\
Mini  
& 94.68 & \textbf{80.16} & \textbf{75.25} & 76.74
& 98.52 & 94.30 & 92.38 & 91.67
& 89.87 & 76.24 & 62.80 & 61.53 \\
Tiny  
& 94.95 & 78.91 & 73.23 & 75.25
& 98.10 & 94.06 & 90.70 & 91.26
& 89.14 & 74.42 & 61.06 & 58.57 \\
Small 
& \textbf{95.41} & \textbf{80.16} & 74.56 & \textbf{76.76}
& \textbf{98.60} & \textbf{94.88} & \textbf{92.70} & \textbf{92.47}
& 88.91 & \textbf{77.52} & \textbf{64.75} & \textbf{64.13} \\
\bottomrule
\end{tabular}
\end{table*}

Table~\ref{tab:scratch_results} shows the results for the four variants. Small performs best overall. It gives the highest accuracy, macro-F1, and $\kappa$ on both Chapman-Shaoxing and PTB-XL, and the highest AUC and $\kappa$ on CPSC2018. The main exception is Mini, which gives a slightly higher macro-F1 than Small on CPSC2018 (75.25 vs.\ 74.56). Micro gives the highest AUC on PTB-XL.

The effect of model size is small and not always in the same direction. Larger models tend to give better results on Chapman-Shaoxing and PTB-XL, but the smaller models are still strong. For example, Micro (0.21M) matches or beats Small (4.17M) on PTB-XL AUC, and Mini (0.78M) is close to Small on all three datasets. Because many of the gaps between variants are small, we treat these differences as a general trend rather than a strict ranking. In practice, this is a useful property, since it shows that the model already works well even with very few parameters.

Based on these results, we use the Small backbone for the self-supervised experiments. It gives the best or joint-best macro-F1 and $\kappa$ on two of the three datasets and stays competitive on CPSC2018, so it is a reliable choice for studying JEPA-style pretraining. Mini is a good lightweight option when a smaller model is needed.

\subsection{Self-Supervised Pretraining and Downstream Transfer}
\label{sec:ssl_transfer}

We next evaluate whether JEPA-style pretraining improves downstream ECG classification. The Small backbone is first pretrained on unlabeled CODE-15\% recordings and then adapted to CPSC2018, Chapman-Shaoxing, and PTB-XL. The purpose of this experiment is to compare the same architecture under three settings: supervised training from scratch, LoRA-based adaptation from the SSL-pretrained encoder, and full fine-tuning from the SSL-pretrained encoder.

\begin{figure*}[!t]
    \centering
    \begin{subfigure}{0.48\textwidth}
        \centering
        \includegraphics[width=\linewidth]{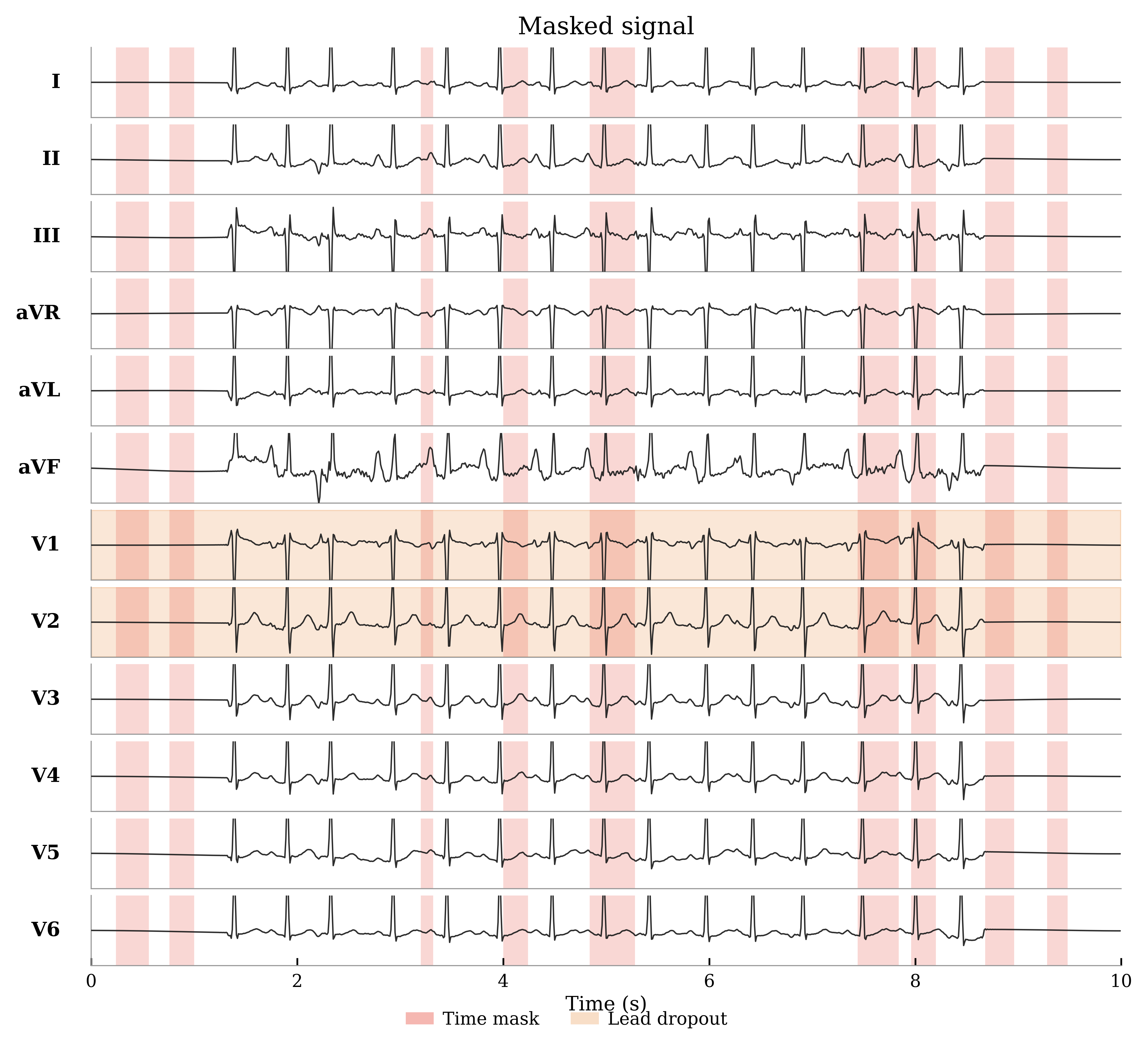}
        \caption{25\% temporal masking.}
        \label{fig:masking_strategy_25}
    \end{subfigure}
    \hfill
    \begin{subfigure}{0.48\textwidth}
        \centering
        \includegraphics[width=\linewidth]{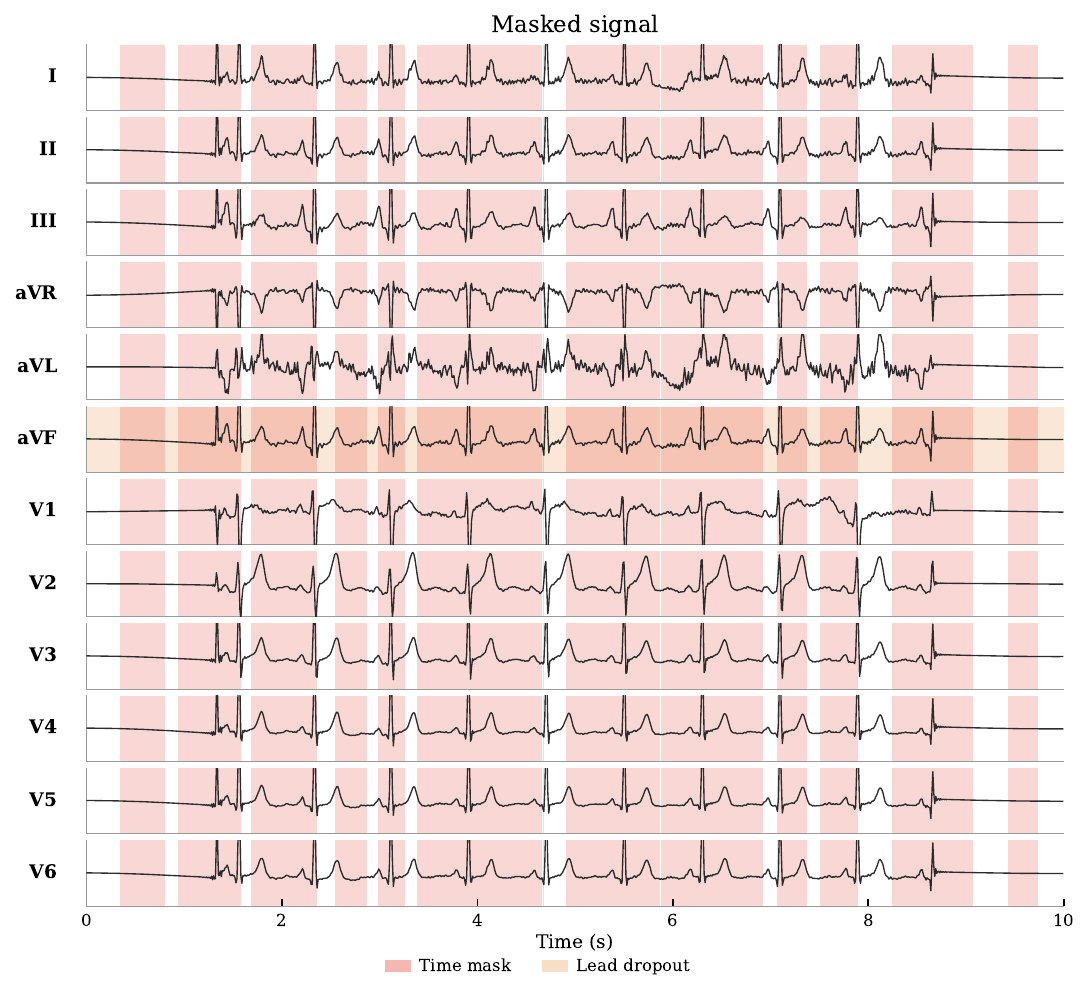}
        \caption{75\% temporal masking.}
        \label{fig:masking_strategy_75}
    \end{subfigure}
    \caption{Representative ECG waveform corruption under different temporal masking ratios. Temporal masks are sampled at the latent token resolution and projected back to the waveform domain before the online encoder. Lower masking preserves more visible waveform morphology, whereas higher masking imposes a stronger contextual prediction task. Lead dropout is applied as an additional channel-level corruption.}
    \label{fig:masking_strategy}
\end{figure*}

Figure~\ref{fig:masking_strategy} shows representative corruption examples under two temporal masking ratios. Random temporal span masks are sampled at the latent temporal resolution and expanded back to the waveform domain, where the corresponding ECG regions are corrupted before entering the online encoder. The 25\% setting preserves more visible waveform morphology, whereas the 75\% setting removes a larger portion of the temporal context and therefore imposes a harder latent prediction task. Lead dropout is also applied as an additional channel-level corruption. In the main transfer experiments, the temporal masking ratio is fixed at 75\%, while the lower masking ratios are evaluated separately in the masking-sensitivity analysis. The target branch receives the corresponding clean ECG and provides latent targets through the momentum encoder; thus, the model predicts latent representations from partially corrupted ECG signals rather than reconstructing raw waveforms.

\begin{table*}[t]
\centering
\caption{Transfer results of the Small backbone on CPSC2018, Chapman-Shaoxing, and PTB-XL. SSL-pretrained models are adapted using LoRA and full fine-tuning. All values are reported in percentage (\%).}
\label{tab:full_label_small_ssl}
\vspace{2pt}
\renewcommand{\arraystretch}{1.05}
\setlength{\tabcolsep}{3.2pt}
\footnotesize
\begin{tabular}{lcccccccccccc}
\toprule
\multirow{2}{*}{\textbf{Mode}}
& \multicolumn{4}{c}{\textbf{CPSC2018}}
& \multicolumn{4}{c}{\textbf{Chapman-Shaoxing}}
& \multicolumn{4}{c}{\textbf{PTB-XL}} \\
\cmidrule(lr){2-5} \cmidrule(lr){6-9} \cmidrule(lr){10-13}
& \textbf{AUC} & \textbf{Acc} & \textbf{F1} & \textbf{$\kappa$}
& \textbf{AUC} & \textbf{Acc} & \textbf{F1} & \textbf{$\kappa$}
& \textbf{AUC} & \textbf{Acc} & \textbf{F1} & \textbf{$\kappa$} \\
\midrule
Scratch  
& 95.41 & 80.16 & 74.56 & 76.76
& 98.60 & 94.88 & 92.70 & 92.47
& 88.91 & 77.52 & \textbf{64.75} & \textbf{64.13} \\
SSL-LoRA 
& 95.39 & 82.03 & 77.58 & 78.93
& \textbf{99.15} & \textbf{97.09} & \textbf{94.92} & \textbf{95.71}
& \textbf{90.21} & 77.03 & 61.56 & 63.05 \\
SSL-Full 
& \textbf{96.67} & \textbf{82.97} & \textbf{78.58} & \textbf{80.03}
& 98.14 & 96.27 & 94.10 & 94.55
& 89.92 & \textbf{77.82} & 61.95 & 63.08 \\
\bottomrule
\end{tabular}
\end{table*}

Table~\ref{tab:full_label_small_ssl} reports the full-label transfer results. The results show that SSL pretraining provides clear gains on CPSC2018 and Chapman-Shaoxing. On CPSC2018, SSL-Full achieves the best performance across all metrics, indicating that the pretrained encoder remains useful even when full labeled data are available. On Chapman-Shaoxing, SSL-LoRA gives the strongest results, suggesting that parameter-efficient adaptation can preserve and exploit the pretrained representation effectively.

On PTB-XL, the effect is more metric-dependent. SSL-LoRA gives the highest AUC and SSL-Full gives the highest accuracy, whereas scratch training remains better in macro-F1 and $\kappa$. This indicates that SSL initialization improves ranking-oriented separability but does not uniformly improve the final class-balanced decision metrics under full supervision. Given the stronger class imbalance in PTB-XL, this behavior suggests that SSL pretraining is beneficial but its impact depends on the downstream label distribution and adaptation mode.

\subsection{Low-Label Transfer Analysis}
\label{sec:low_label}

We further evaluate downstream transfer under limited supervision. For each dataset, we sample stratified subsets containing 1\%, 5\%, 10\%, and 20\% of the labeled training data. The same subset manifest is used across scratch, SSL-LoRA, and SSL-Full to ensure matched comparisons. Each experiment is repeated over three sampling seeds, and results are reported as mean $\pm$ standard deviation.

\begin{table*}[t]
\centering
\caption{Low-label transfer results on CPSC2018 using different fractions of the labeled training data. SSL-pretrained models are adapted using LoRA and full fine-tuning. Results are reported as mean $\pm$ standard deviation over three runs, in percentage (\%).}
\label{tab:low_label_cpsc2018}
\renewcommand{\arraystretch}{1.0}
\setlength{\tabcolsep}{4pt}
\footnotesize
\begin{tabular}{lccccc}
\toprule
\textbf{Train Fraction} & \textbf{Mode} & \textbf{AUC} & \textbf{Acc} & \textbf{Macro-F1} & \textbf{$\kappa$} \\
\midrule
\multirow{3}{*}{1\%}
& Scratch  & 71.62 $\pm$ 1.24 & 36.98 $\pm$ 2.13 & 28.11 $\pm$ 2.90 & 26.35 $\pm$ 2.68 \\
& SSL-LoRA & 81.37 $\pm$ 1.14 & 51.15 $\pm$ 3.56 & 39.58 $\pm$ 6.31 & 42.43 $\pm$ 4.17 \\
& SSL-Full & \textbf{82.92 $\pm$ 1.00} & \textbf{54.53 $\pm$ 3.17} & \textbf{45.70 $\pm$ 4.33} & \textbf{46.60 $\pm$ 3.74} \\
\midrule
\multirow{3}{*}{5\%}
& Scratch  & 84.41 $\pm$ 0.52 & 57.03 $\pm$ 1.84 & 49.50 $\pm$ 2.73 & 49.84 $\pm$ 2.19 \\
& SSL-LoRA & \textbf{92.06 $\pm$ 0.27} & \textbf{72.45 $\pm$ 0.72} & \textbf{65.30 $\pm$ 2.40} & \textbf{67.71 $\pm$ 1.00} \\
& SSL-Full & 91.65 $\pm$ 0.50 & 71.82 $\pm$ 0.89 & 65.12 $\pm$ 2.08 & 66.99 $\pm$ 1.15 \\
\midrule
\multirow{3}{*}{10\%}
& Scratch  & 88.74 $\pm$ 0.30 & 65.57 $\pm$ 1.41 & 58.44 $\pm$ 2.52 & 59.73 $\pm$ 1.64 \\
& SSL-LoRA & 92.96 $\pm$ 0.08 & 74.90 $\pm$ 0.18 & 68.00 $\pm$ 0.84 & 70.58 $\pm$ 0.13 \\
& SSL-Full & \textbf{93.56 $\pm$ 0.40} & \textbf{75.73 $\pm$ 0.33} & \textbf{70.53 $\pm$ 0.92} & \textbf{71.62 $\pm$ 0.33} \\
\midrule
\multirow{3}{*}{20\%}
& Scratch  & 91.66 $\pm$ 0.32 & 70.99 $\pm$ 0.45 & 64.67 $\pm$ 0.33 & 65.99 $\pm$ 0.47 \\
& SSL-LoRA & \textbf{94.73 $\pm$ 0.40} & 78.91 $\pm$ 0.00 & 73.94 $\pm$ 0.77 & 75.30 $\pm$ 0.02 \\
& SSL-Full & 94.58 $\pm$ 0.32 & \textbf{79.22 $\pm$ 1.65} & \textbf{74.72 $\pm$ 1.58} & \textbf{75.68 $\pm$ 1.90} \\
\bottomrule
\end{tabular}
\end{table*}
\begin{table*}[t]
\centering
\caption{Low-label transfer results on Chapman-Shaoxing using different fractions of the labeled training data. SSL-pretrained models are adapted using LoRA and full fine-tuning. Results are reported as mean $\pm$ standard deviation over three runs, in percentage (\%).}
\label{tab:low_label_chapman}
\renewcommand{\arraystretch}{1.0}
\setlength{\tabcolsep}{4pt}
\footnotesize
\begin{tabular}{lccccc}
\toprule
\textbf{Train Fraction} & \textbf{Mode} & \textbf{AUC} & \textbf{Acc} & \textbf{Macro-F1} & \textbf{$\kappa$} \\
\midrule
\multirow{3}{*}{1\%}
& Scratch  & 85.05 $\pm$ 1.56 & 66.67 $\pm$ 1.34 & 59.20 $\pm$ 1.86 & 50.03 $\pm$ 2.04 \\
& SSL-LoRA & \textbf{98.27 $\pm$ 0.17} & \textbf{90.84 $\pm$ 0.87} & \textbf{86.99 $\pm$ 1.72} & \textbf{86.49 $\pm$ 1.33} \\
& SSL-Full & 98.15 $\pm$ 0.29 & 90.65 $\pm$ 1.05 & 86.61 $\pm$ 1.40 & 86.22 $\pm$ 1.57 \\
\midrule
\multirow{3}{*}{5\%}
& Scratch  & 92.51 $\pm$ 0.95 & 80.75 $\pm$ 3.54 & 76.09 $\pm$ 4.02 & 71.42 $\pm$ 5.78 \\
& SSL-LoRA & 98.30 $\pm$ 0.22 & 91.73 $\pm$ 0.51 & 88.41 $\pm$ 0.69 & 87.92 $\pm$ 0.75 \\
& SSL-Full & \textbf{98.37 $\pm$ 0.18} & \textbf{91.93 $\pm$ 0.64} & \textbf{88.78 $\pm$ 0.96} & \textbf{88.24 $\pm$ 0.95} \\
\midrule
\multirow{3}{*}{10\%}
& Scratch  & 95.34 $\pm$ 0.04 & 86.81 $\pm$ 0.27 & 82.43 $\pm$ 1.36 & 80.39 $\pm$ 0.42 \\
& SSL-LoRA & \textbf{98.38 $\pm$ 0.60} & 92.28 $\pm$ 1.57 & 89.48 $\pm$ 1.62 & 88.77 $\pm$ 2.23 \\
& SSL-Full & 97.98 $\pm$ 0.33 & \textbf{93.17 $\pm$ 0.34} & \textbf{90.00 $\pm$ 0.53} & \textbf{89.97 $\pm$ 0.47} \\
\midrule
\multirow{3}{*}{20\%}
& Scratch  & 97.05 $\pm$ 0.54 & 90.26 $\pm$ 1.17 & 87.13 $\pm$ 1.06 & 85.69 $\pm$ 1.72 \\
& SSL-LoRA & \textbf{98.15 $\pm$ 0.74} & 93.36 $\pm$ 1.15 & 90.37 $\pm$ 1.40 & 90.28 $\pm$ 1.68 \\
& SSL-Full & 98.07 $\pm$ 0.43 & \textbf{93.75 $\pm$ 0.29} & \textbf{90.78 $\pm$ 0.60} & \textbf{90.85 $\pm$ 0.41} \\
\bottomrule
\end{tabular}
\end{table*}

\begin{table*}[t]
\centering
\caption{Low-label transfer results on PTB-XL using different fractions of the labeled training data. SSL-pretrained models are adapted using LoRA and full fine-tuning. Results are reported as mean $\pm$ standard deviation over three runs, in percentage (\%).}
\label{tab:low_label_ptbxl}
\renewcommand{\arraystretch}{1.0}
\setlength{\tabcolsep}{4pt}
\footnotesize
\begin{tabular}{lccccc}
\toprule
\textbf{Train Fraction} & \textbf{Mode} & \textbf{AUC} & \textbf{Acc} & \textbf{Macro-F1} & \textbf{$\kappa$} \\
\midrule
\multirow{3}{*}{1\%}
& Scratch  & 74.51 $\pm$ 0.81 & 62.53 $\pm$ 0.73 & 43.04 $\pm$ 1.27 & 39.77 $\pm$ 0.67 \\
& SSL-LoRA & 75.88 $\pm$ 0.80 & 64.83 $\pm$ 2.43 & 43.36 $\pm$ 1.15 & 41.40 $\pm$ 2.41 \\
& SSL-Full & \textbf{76.51 $\pm$ 1.51} & \textbf{65.27 $\pm$ 2.43} & \textbf{45.45 $\pm$ 0.65} & \textbf{43.29 $\pm$ 1.89} \\
\midrule
\multirow{3}{*}{5\%}
& Scratch  & 81.65 $\pm$ 1.19 & 70.08 $\pm$ 1.91 & 50.96 $\pm$ 0.83 & 50.74 $\pm$ 2.49 \\
& SSL-LoRA & 82.51 $\pm$ 0.57 & \textbf{70.79 $\pm$ 1.36} & 52.11 $\pm$ 1.46 & 51.78 $\pm$ 1.38 \\
& SSL-Full & \textbf{82.78 $\pm$ 1.32} & 70.46 $\pm$ 1.36 & \textbf{52.64 $\pm$ 1.52} & \textbf{52.06 $\pm$ 1.09} \\
\midrule
\multirow{3}{*}{10\%}
& Scratch  & 83.08 $\pm$ 1.11 & 70.34 $\pm$ 1.82 & 55.36 $\pm$ 1.18 & 53.28 $\pm$ 2.48 \\
& SSL-LoRA & 84.11 $\pm$ 1.34 & 71.49 $\pm$ 2.19 & 53.89 $\pm$ 1.41 & \textbf{54.03 $\pm$ 1.72} \\
& SSL-Full & \textbf{84.49 $\pm$ 1.38} & \textbf{71.90 $\pm$ 1.34} & \textbf{55.43 $\pm$ 1.18} & 53.98 $\pm$ 1.02 \\
\midrule
\multirow{3}{*}{20\%}
& Scratch  & 85.86 $\pm$ 1.00 & 73.74 $\pm$ 1.05 & 57.43 $\pm$ 0.47 & \textbf{57.60 $\pm$ 1.73} \\
& SSL-LoRA & 86.41 $\pm$ 1.36 & 73.05 $\pm$ 1.83 & \textbf{57.79 $\pm$ 0.89} & 56.54 $\pm$ 2.03 \\
& SSL-Full & \textbf{86.90 $\pm$ 1.02} & \textbf{73.78 $\pm$ 0.43} & 57.15 $\pm$ 0.83 & 57.09 $\pm$ 0.27 \\
\bottomrule
\end{tabular}
\end{table*}

Tables~\ref{tab:low_label_cpsc2018}--\ref{tab:low_label_ptbxl} report the reduced-label transfer results. In general, SSL initialization provides the largest gains when the amount of labeled data is small. The effect is most pronounced on CPSC2018 and Chapman-Shaoxing, where SSL-based adaptation consistently improves over scratch across all training fractions. On CPSC2018, full fine-tuning gives the strongest overall performance at most fractions, while LoRA remains competitive and gives the highest AUC at selected fractions. On Chapman-Shaoxing, LoRA is particularly effective at the lowest label fraction, whereas full fine-tuning becomes slightly stronger as more labels are available.

The trend on PTB-XL is more moderate. SSL-based adaptation improves most metrics at 1\%, 5\%, and 10\%, indicating that the pretrained encoder provides a useful representation prior under limited supervision. At 20\%, the differences become smaller and metric-dependent, with scratch remaining competitive for $\kappa$ while SSL-based models retain advantages in AUC, accuracy, or macro-F1. This behavior is consistent with the stronger class imbalance of PTB-XL and suggests that the benefit of SSL initialization depends on both the label budget and the downstream label distribution.

Overall, these results show that the JEPA-style pretraining is most beneficial in low-label regimes. The gains are especially clear when only a small portion of the training set is available, supporting the role of the pretrained encoder as a transferable initialization for data-efficient ECG classification.

\FloatBarrier

\subsection{Sensitivity to Masking Ratio}
\label{sec:mask_sensitivity}

We further examine the effect of the temporal masking ratio used during JEPA-style pretraining. In the main experiments, the masking ratio is fixed at 75\%, which defines the default pretraining configuration used for the reported SSL transfer results. To assess the influence of corruption strength, we additionally pretrain the Small backbone using lower masking ratios of 25\% and 50\%. The downstream evaluation is conducted using full fine-tuning on CPSC2018, Chapman-Shaoxing, and PTB-XL, while keeping the backbone, pretraining data, downstream splits, optimizer, and adaptation protocol unchanged. To isolate the effect of the masking ratio, this analysis uses full fine-tuning for all masking configurations; LoRA-based adaptation is evaluated separately under the default 75\% masking configuration in Table~\ref{tab:full_label_small_ssl}.

\begin{table*}[t]
\centering
\caption{Sensitivity of JEPA-style pretraining to the temporal masking ratio using the Small backbone and full fine-tuning. The 75\% setting corresponds to the default configuration used in the main experiments. All values are reported in percentage (\%).}
\label{tab:mask_ratio_sensitivity}
\vspace{2pt}
\renewcommand{\arraystretch}{1.05}
\setlength{\tabcolsep}{3.2pt}
\footnotesize
\begin{tabular}{lcccccccccccc}
\toprule
\multirow{2}{*}{\textbf{Mask Ratio}}
& \multicolumn{4}{c}{\textbf{CPSC2018}}
& \multicolumn{4}{c}{\textbf{Chapman-Shaoxing}}
& \multicolumn{4}{c}{\textbf{PTB-XL}} \\
\cmidrule(lr){2-5} \cmidrule(lr){6-9} \cmidrule(lr){10-13}
& \textbf{AUC} & \textbf{Acc} & \textbf{F1} & \textbf{$\kappa$}
& \textbf{AUC} & \textbf{Acc} & \textbf{F1} & \textbf{$\kappa$}
& \textbf{AUC} & \textbf{Acc} & \textbf{F1} & \textbf{$\kappa$} \\
\midrule
25\%
& 94.32 & 76.88 & 71.13 & 72.90
& 97.61 & 96.16 & 93.69 & 94.37
& 89.50 & 77.64 & 63.25 & \textbf{63.58} \\
50\%
& 96.30 & \textbf{83.13} & \textbf{79.19} & \textbf{80.26}
& 98.57 & 96.16 & \textbf{94.29} & 94.35
& 89.42 & 76.73 & \textbf{63.69} & 62.73 \\
75\% (default)
& \textbf{96.67} & 82.97 & 78.58 & 80.03
& \textbf{98.14} & \textbf{96.27} & 94.10 & \textbf{94.55}
& \textbf{89.92} & \textbf{77.82} & 61.95 & 63.08 \\
\bottomrule
\end{tabular}
\end{table*}

Table~\ref{tab:mask_ratio_sensitivity} reports the masking-ratio sensitivity results. The effect of the masking ratio is not monotonic across datasets or metrics. On CPSC2018, the 50\% masking ratio gives the strongest decision-level performance, achieving 83.13\% accuracy, 79.19\% macro-F1, and 80.26\% $\kappa$. The default 75\% setting remains slightly higher in AUC, but the 50\% setting provides better class-balanced and agreement-based performance. This suggests that an intermediate corruption level can provide a favorable balance between preserving discriminative waveform morphology and enforcing contextual latent prediction.

On Chapman-Shaoxing, the three masking ratios produce relatively close results. The 50\% setting gives the highest macro-F1, reaching 94.29\%, while the default 75\% setting gives the highest accuracy and $\kappa$. The 25\% setting is also competitive, but remains slightly below the 50\% and 75\% configurations in the main decision metrics. These results indicate that the proposed JEPA-style objective is stable on this rhythm-classification dataset, with only modest variation across masking ratios.

On PTB-XL, the trend differs from CPSC2018 and Chapman-Shaoxing. The 50\% masking ratio obtains the highest macro-F1, whereas the 75\% setting gives the best AUC and accuracy, and the 25\% setting gives the highest $\kappa$. This metric-dependent behavior is consistent with the stronger class imbalance and diagnostic heterogeneity of PTB-XL. In particular, lower or intermediate masking ratios may better preserve morphology-sensitive diagnostic cues, while stronger masking can improve overall separability and accuracy.

Overall, the sensitivity analysis shows that the masking ratio influences the balance between morphology preservation and contextual prediction difficulty. The 75\% configuration remains a strong and consistent default, especially in terms of AUC and overall accuracy, and is therefore retained for the main transfer experiments. At the same time, the 50\% results show that an intermediate corruption level can improve macro-F1 on all three datasets, suggesting that masking strength is an important design factor for ECG latent predictive pretraining.

\subsection{Qualitative Interpretability Analysis}
\label{sec:qualitative_interpretability}
To complement the quantitative results, we perform a simple qualitative analysis of the Small backbone. The aim is not to explain individual decisions clinically, but to check whether the model responds to meaningful parts of the ECG waveform. We use two attribution methods. Grad-CAM++~\cite{chattopadhay2018gradcampp} is applied to the Stage~2 representation, the last temporally localized feature map before the global self-attention stage, giving a class-specific view of the time regions that contribute to the prediction. Attention rollout~\cite{abnar2020rollout} is then computed over the Stage~3 self-attention blocks to show how information is combined in the final stage. Since the two maps capture different aspects, class-specific attribution and attention-based information flow, we show them together for each example.

\begin{figure*}[t]
  \centering

  \begin{subfigure}[t]{0.315\textwidth}
      \centering
      \includegraphics[width=\linewidth]{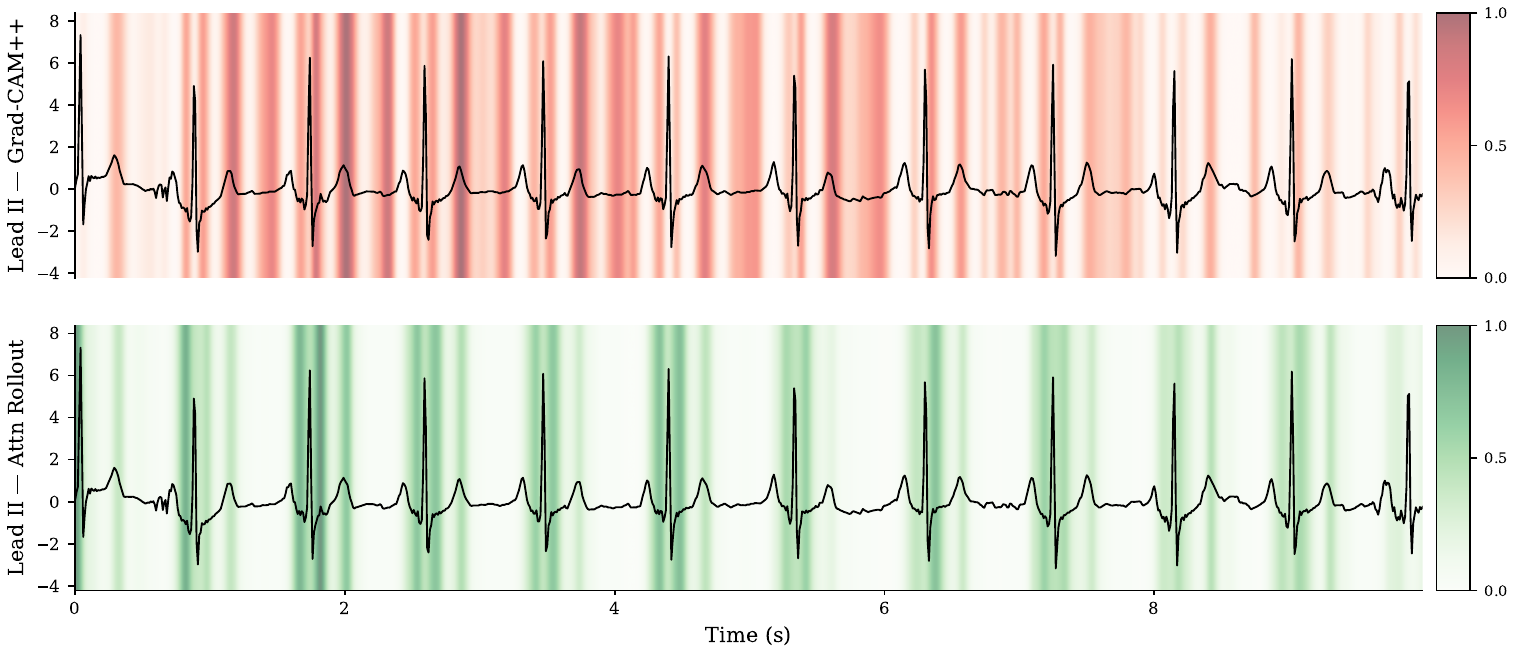}
      \caption{Correct NORM prediction on PTB-XL.}
      \label{fig:viz_ptbxl_correct}
  \end{subfigure}
  \hfill
  \begin{subfigure}[t]{0.315\textwidth}
      \centering
      \includegraphics[width=\linewidth]{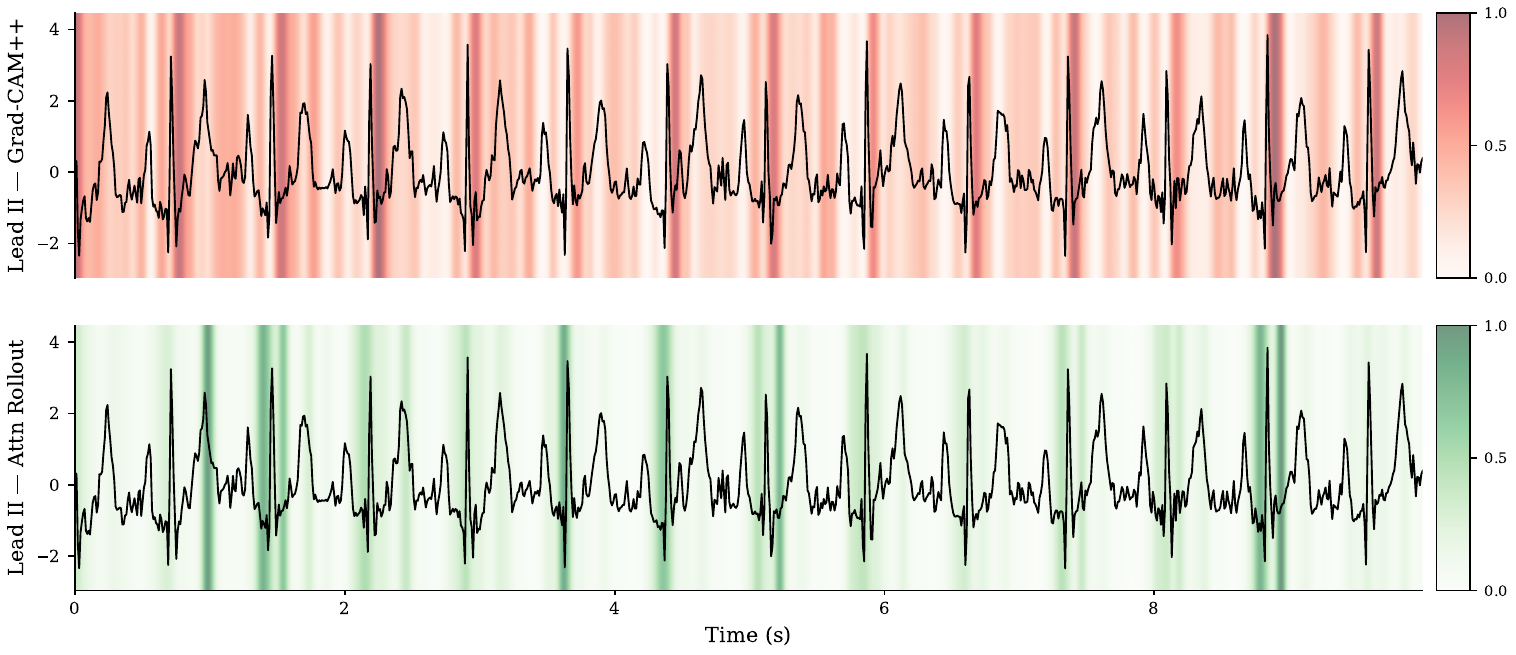}
      \caption{NORM misclassified as MI on PTB-XL.}
      \label{fig:viz_ptbxl_wrong}
  \end{subfigure}
  \hfill
  \begin{subfigure}[t]{0.315\textwidth}
      \centering
      \includegraphics[width=\linewidth]{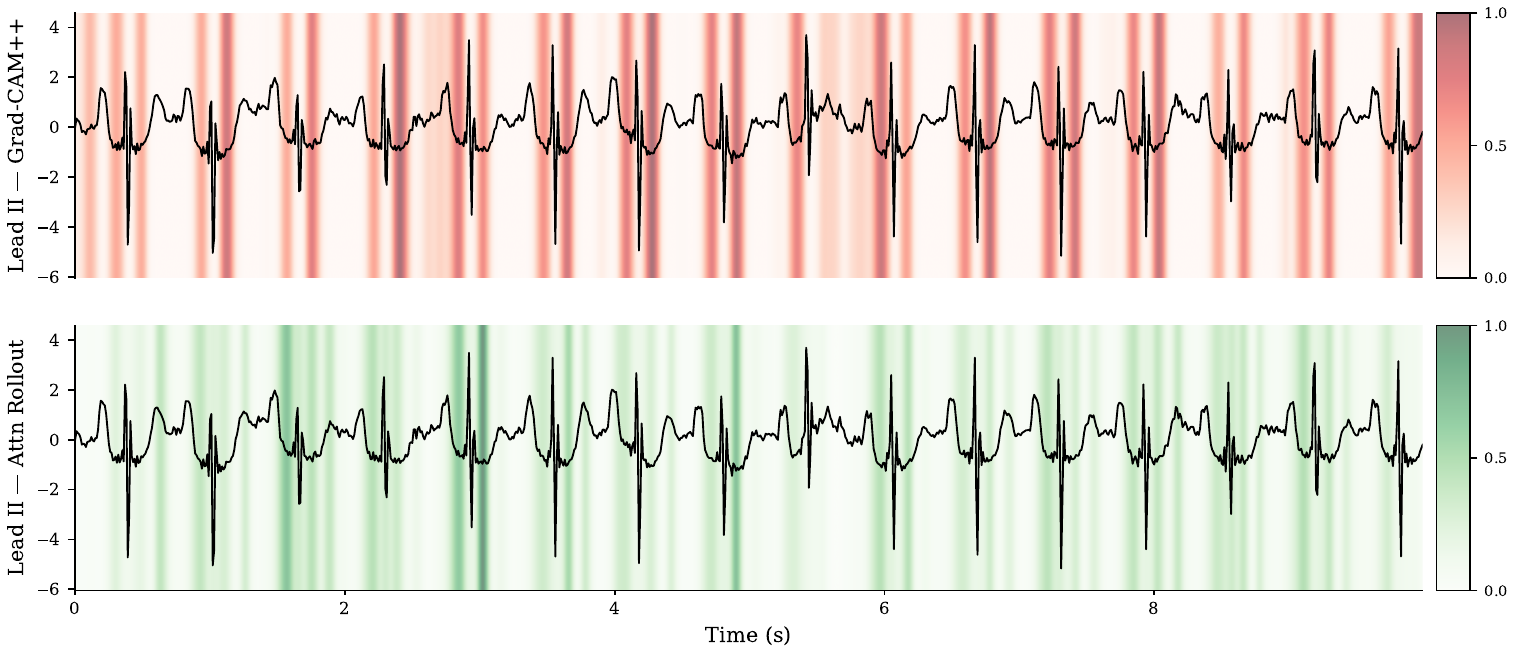}
      \caption{Normal misclassified as IAVB on CPSC2018.}
      \label{fig:viz_cpsc_wrong}
  \end{subfigure}

  \caption{Qualitative interpretation of the Small backbone. 
  Panels (a) and (b) show correct and incorrect predictions on PTB-XL, respectively, while panel (c) shows a misclassification on CPSC2018. 
  In each panel, the upper strip presents the Grad-CAM++ map~\cite{chattopadhay2018gradcampp} computed from the Stage~2 representation, and the lower strip presents the attention-rollout map~\cite{abnar2020rollout} aggregated from the Stage~3 self-attention blocks.}
  \label{fig:qualitative_viz}
\end{figure*}

Figure~\ref{fig:qualitative_viz} shows representative correct and incorrect predictions. In the correct PTB-XL NORM example (Fig.~\ref{fig:viz_ptbxl_correct}), both maps mainly highlight regions around the repeated beats, in particular near the QRS complexes. In the NORM$\rightarrow$MI case (Fig.~\ref{fig:viz_ptbxl_wrong}), the highlighted regions shift toward parts of the signal involving the ST segment and T-wave. This does not confirm that the prediction is clinically correct, but it suggests that the error is linked to waveform regions plausibly related to the confused class. A similar pattern appears in the CPSC2018 Normal$\rightarrow$IAVB case (Fig.~\ref{fig:viz_cpsc_wrong}), where the maps concentrate around beat-level structures rather than unrelated parts of the signal.

Overall, these examples suggest that the model tends to respond to structured regions of the ECG and that its errors are often linked to less regular or ambiguous segments. We treat this as qualitative support only, complementing the classification metrics rather than providing a clinical interpretation.

\subsection{Comparison with State-of-the-Art Methods}
\label{sec:sota_comparison}

We compare the proposed hybrid CNN--SSM--Attention model with recent 12-lead ECG classification methods. The comparison is organized into three parts. First, we evaluate the reduced-label setting on CPSC2018 and PTB-XL using only 20\% of the labeled training data, which is particularly useful for assessing the transfer value of self-supervised pretraining. Second, we report full-label results on CPSC2018 and PTB-XL under the fixed-split protocol used in the main experiments. Third, we present a separate comparison on Chapman-Shaoxing. Throughout, we also include 10-fold cross-validation results for our variants as an additional robustness check. These 10-fold results are not intended to replace the fixed-split comparison with prior work, but to verify that the observed trends are not tied to a single data split.

Table~\ref{tab:sota_20percent} reports the comparison at 20\% labeled training data. Two main observations can be made. First, even without self-supervised initialization, the proposed backbone provides a strong supervised baseline. The random-initialized variant improves over the random-initialized baseline of Ma et al. on both datasets and remains competitive with several self-supervised baselines. This supports the effectiveness of the proposed architecture itself. Second, JEPA-style pretraining provides clear gains in the low-label regime, especially on CPSC2018. On this dataset, SSL-Full achieves the best accuracy, macro-F1, and $\kappa$, while SSL-LoRA obtains the best AUC. These results show that the pretrained representation transfers effectively when labeled supervision is limited.

On PTB-XL, the improvement from pretraining is more moderate. The proposed variants remain competitive with the matched baselines, but the gains are metric-dependent. SSL-LoRA gives the highest macro-F1 among the compared methods, whereas the random-initialized backbone gives the highest $\kappa$. This suggests that, for PTB-XL, the proposed backbone already provides a strong supervised representation, while JEPA-style pretraining adds a smaller and less uniform benefit than on CPSC2018.

\begin{table*}[t]
\centering
\caption{Comparison at 20\% labeled training data on CPSC2018 and PTB-XL. All values are reported in percentage (\%). Bold values indicate the best result within each dataset and metric.}
\label{tab:sota_20percent}
\renewcommand{\arraystretch}{1.05}
\setlength{\tabcolsep}{2.6pt}
\scriptsize
\resizebox{\textwidth}{!}{%
\begin{tabular}{llcccccccc}
\toprule
\multirow{2}{*}{\textbf{Method}}
& \multirow{2}{*}{\textbf{Setting}}
& \multicolumn{4}{c}{\textbf{CPSC2018}}
& \multicolumn{4}{c}{\textbf{PTB-XL}} \\
\cmidrule(lr){3-6}\cmidrule(lr){7-10}
& & \textbf{AUC} & \textbf{Acc} & \textbf{F1} & \textbf{$\kappa$}
  & \textbf{AUC} & \textbf{Acc} & \textbf{F1} & \textbf{$\kappa$} \\
\midrule
Ma et al. (2026) & Random Init.
& 86.95 & 68.03 & 58.21 & 61.53
& 85.76 & 72.37 & 54.31 & 53.50 \\
Ma et al. (2026) & MoCo
& 80.30 & 51.04 & 41.50 & 41.75
& 79.70 & 65.93 & 45.21 & 43.25 \\
Ma et al. (2026) & NNCLR
& 86.86 & 64.36 & 54.98 & 58.00
& 85.76 & \textbf{74.39} & 55.75 & 56.38 \\
Ma et al. (2026) & SimCLR
& 89.52 & 68.84 & 60.82 & 63.32
& 86.64 & 73.97 & 56.07 & 56.13 \\
Ma et al. (2026) & DCCLR
& 88.40 & 67.39 & 59.57 & 61.61
& 87.21 & 74.24 & 56.10 & 56.67 \\
Ma et al. (2026) & SimSiam-GL
& 88.41 & 69.06 & 60.63 & 63.53
& \textbf{87.44} & 74.04 & 56.99 & 56.06 \\
\midrule
Ours & Random Init.
& 91.66 & 70.99 & 64.67 & 65.99
& 85.86 & 73.74 & 57.43 & \textbf{57.60} \\
Ours & SSL-LoRA
& \textbf{94.73} & 78.91 & 73.94 & 75.30
& 86.41 & 73.05 & \textbf{57.79} & 56.54 \\
Ours & SSL-Full
& 94.58 & \textbf{79.22} & \textbf{74.72} & \textbf{75.68}
& 86.90 & 73.78 & 57.15 & 57.09 \\
\bottomrule
\end{tabular}%
}
\end{table*}

Table~\ref{tab:sota_fulllabel} reports the full-label comparison. Under the fixed-split protocol, the proposed method achieves the strongest overall performance on CPSC2018. SSL-Full obtains the best AUC, accuracy, macro-F1, and $\kappa$, showing that JEPA-style initialization remains useful even when the full labeled training set is available. The random-initialized backbone is also strong, which further supports the contribution of the proposed hybrid architecture.

The 10-fold cross-validation results provide additional evidence for the stability of these trends on CPSC2018. SSL-LoRA gives the highest accuracy, macro-F1, and $\kappa$, while SSL-Full gives the highest AUC. This indicates that both full fine-tuning and parameter-efficient adaptation can benefit from the pretrained representation, and that the advantage is not limited to one fixed split.

On PTB-XL, the comparison is more nuanced. Prior task-specific self-supervised methods achieve the highest fixed-split AUC and accuracy, while our random-initialized backbone achieves the strongest macro-F1 and $\kappa$ among the fixed-split results reported in the table. In the 10-fold setting, SSL-LoRA gives the highest AUC and accuracy, and a marginally higher $\kappa$, while the random-initialized model gives the highest macro-F1. These results suggest that, on PTB-XL, JEPA-style pretraining can improve ranking-oriented or overall metrics, but it does not consistently improve class-balanced decision metrics after adaptation. This behavior is likely related to the stronger class imbalance of PTB-XL.

\begin{table*}[t]
\centering
\caption{Full-label results and comparison with recent ECG classification methods on CPSC2018 and PTB-XL. All values are reported in percentage (\%).}
\label{tab:sota_fulllabel}
\renewcommand{\arraystretch}{1.05}
\setlength{\tabcolsep}{2.4pt}
\scriptsize
\resizebox{\textwidth}{!}{%
\begin{tabular}{llcccccccc}
\toprule
\multirow{2}{*}{\textbf{Method}}
& \multirow{2}{*}{\textbf{Setting}}
& \multicolumn{4}{c}{\textbf{CPSC2018}}
& \multicolumn{4}{c}{\textbf{PTB-XL}} \\
\cmidrule(lr){3-6}\cmidrule(lr){7-10}
&  & \textbf{AUC} & \textbf{Acc} & \textbf{F1} & \textbf{$\kappa$}
   & \textbf{AUC} & \textbf{Acc} & \textbf{F1} & \textbf{$\kappa$} \\
\midrule
Liu et al. (2023) & Frozen
& 92.01 & 68.07 & 63.22 & --
& 86.76 & 76.00 & 57.27 & -- \\
Shi et al. (2024) & SSL-Full
& 95.38 & 74.40 & -- & --
& \textbf{91.23} & \textbf{78.87} & -- & -- \\
Liu et al. (2025) & SSL-Full
& 93.98 & -- & 68.82 & --
& 90.46 & -- & 61.69 & -- \\
Ma et al. (2026) & Proposed
& 92.53 & 77.47 & 69.99 & --
& 90.23 & 76.86 & 63.11 & -- \\
\midrule
Ours & Random Init.
& 95.41 & 80.16 & 74.56 & 76.76
& 88.91 & 77.52 & \textbf{64.75} & \textbf{64.13} \\
Ours & SSL-LoRA
& 95.39 & 82.03 & 77.58 & 78.93
& 90.21 & 77.03 & 61.56 & 63.05 \\
Ours & SSL-Full
& \textbf{96.67} & \textbf{82.97} & \textbf{78.58} & \textbf{80.03}
& 89.92 & 77.82 & 61.95 & 63.08 \\
\midrule
Ours & Random Init. / 10-CV
& 95.04$\pm$1.08 & 80.72$\pm$1.90 & 76.59$\pm$2.27 & 77.42$\pm$2.25
& 91.50$\pm$0.93 & 79.33$\pm$1.15 & \textbf{68.19$\pm$0.76} & 67.20$\pm$1.53 \\
Ours & SSL-LoRA / 10-CV
& 96.08$\pm$1.00 & \textbf{82.81$\pm$1.73} & \textbf{78.96$\pm$2.12} & \textbf{79.87$\pm$2.02}
& \textbf{91.69$\pm$0.54} & \textbf{79.36$\pm$0.98} & 67.55$\pm$1.42 & \textbf{67.29$\pm$1.35} \\
Ours & SSL-Full / 10-CV
& \textbf{96.24$\pm$0.70} & 81.21$\pm$1.87 & 77.01$\pm$2.11 & 77.99$\pm$2.19
& 90.78$\pm$0.37 & 78.52$\pm$0.85 & 64.69$\pm$1.49 & 65.23$\pm$1.26 \\
\bottomrule
\end{tabular}%
}
\end{table*}

Overall, the results show that the proposed backbone is competitive with recent ECG classification methods across different supervision levels and evaluation protocols. JEPA-style pretraining provides the clearest benefit in the reduced-label setting and on CPSC2018, while its effect on PTB-XL is more dependent on the metric and adaptation strategy. This suggests that the proposed architecture can serve as a strong supervised baseline and as an effective platform for further ECG-specific self-supervised learning.

For Chapman-Shaoxing, we report a separate comparison under the four-class rhythm setting of AFIB, GSVT, SB, and SR~\cite{zheng2020chapman}. We use 10-fold cross-validation and report the mean and standard deviation across folds. As a reference, we include the validation baseline reported by Zheng et al.~\cite{zheng2020chapman}, which uses an XGBoost classifier with age, gender, and 230 handcrafted ECG features. In contrast, our model operates directly on the raw 12-lead ECG waveform without demographic variables or engineered clinical descriptors.

\begin{table}[t]
\centering
\caption{Comparison on Chapman-Shaoxing under the four-class rhythm.}
\label{tab:sota_chapman}
\vspace{2pt}
\renewcommand{\arraystretch}{1.05}
\setlength{\tabcolsep}{4.5pt}
\footnotesize
\begin{tabular}{lcc}
\toprule
\textbf{Method} & \textbf{Acc} & \textbf{Macro-F1} \\
\midrule
Zheng et al.~\cite{zheng2020chapman} 
(XGBoost + demographics + handcrafted features)
& \textbf{97.00} & \textbf{96.50} \\
\midrule
Ours (Random Init.) & 94.57$\pm$0.59 & 93.86$\pm$0.72 \\
Ours (SSL-Full)     & \underline{96.34$\pm$0.87} & 95.24$\pm$0.85 \\
Ours (SSL-LoRA)     & 96.15$\pm$0.71 & \underline{95.63$\pm$0.80} \\
\bottomrule
\end{tabular}
\end{table}

Table~\ref{tab:sota_chapman} shows that the proposed model is competitive on Chapman-Shaoxing when trained directly from raw multilead ECG signals. JEPA-style pretraining improves over random initialization for both adaptation modes. Among our variants, SSL-Full achieves the highest accuracy, while SSL-LoRA achieves the highest macro-F1. Although the feature-based baseline of Zheng et al.~\cite{zheng2020chapman} remains stronger overall, the gap is relatively small considering that their method uses demographic information and handcrafted ECG descriptors, whereas our method uses only raw ECG waveforms. These results indicate that the proposed backbone learns useful rhythm representations and benefits from self-supervised initialization on this benchmark.

\section{Conclusion}
\label{sec:conclusion}
In this work, we presented a scalable hybrid CNN--SSM--Attention framework for 12-lead ECG classification, combining convolutional tokenization for local waveform morphology, state-space mixing for efficient temporal modeling, and late self-attention for global token interaction after temporal reduction. We further introduced an ECG-oriented JEPA-style pretraining strategy that predicts latent targets from a momentum encoder instead of reconstructing raw waveforms, applying ECG-specific corruption by projecting latent masks back to the waveform domain. Experiments on CPSC2018, Chapman-Shaoxing, and PTB-XL show that the backbone is a strong supervised baseline even from random initialization and at a compact parameter budget, and that JEPA-style pretraining further improves transfer, most clearly under reduced-label supervision, while LoRA-based adaptation reaches similar gains with far fewer trainable parameters.

Beyond these results, the framework opens several directions for future work. First, it can be pretrained on larger and more diverse ECG datasets to improve generalization. Second, the masking strategy can be refined using lead-aware or rhythm-aware corruption. Third, the same backbone can be extended to other ECG tasks, such as rhythm analysis, abnormality localization, and clinical risk prediction. Together, these directions suggest that efficient sequence modeling combined with ECG-specific predictive pretraining is a promising path toward more transferable 12-lead ECG analysis.

\section*{Data availability}
The downstream datasets used in this study are publicly available. The CODE-15\%, PTB-XL, and CPSC2018 datasets can be accessed from their respective public repositories. Code and trained models will be made available upon reasonable request.

\section*{Funding}
This work was supported by the Ongoing Research Funding Program King Saud University, Riyadh, Saudi Arabia (ORF-2026-1606).

\bibliographystyle{elsarticle-num}
\bibliography{references_bspc}

\end{document}